\documentclass[10pt,twocolumn,letterpaper]{article}

\usepackage[pagenumbers]{cvpr}

\usepackage{bm}

\def\vg{{\bm{g}}}
\def\vx{{\bm{x}}}
\def\vy{{\bm{y}}}

\newcommand{\sg}[1]{\mathrm{sg}\!\left[#1\right]}
\newcommand{\Tmean}{\mathcal{M}}
\newcommand{\dev}{\boldsymbol{\delta}}
\newcommand{\xhat}{\hat{\vx}_0}
\newcommand{\xbar}{\bar{\vx}_0}
\newcommand{\sreal}{s_{\phi}}
\newcommand{\sfake}{s_{\psi}}
\newcommand{\relu}[1]{\left[#1\right]_{+}}
\newcommand{\method}{\textsc{TSD}}
\newcommand{\methodB}{\mbox{\textsc{TSD-B}}}
\newcommand{\methodD}{\mbox{\textsc{TSD-D}}}
\newcommand{\vbone}{\mbox{VBench-1}}
\newcommand{\vbtwo}{\mbox{VBench-2}}

\definecolor{cvprblue}{rgb}{0.21,0.49,0.74}
\usepackage[pagebackref,breaklinks,colorlinks,allcolors=cvprblue]{hyperref}

\def\confName{CVPR}
\def\confYear{2027}

\title{TSD: Few-Step Video Diffusion with Temporal Subspace Distillation}

\author{Ziyi Yang$^{1}$ \quad Yi-Hua Huang$^{1}$ \quad Chirui Chang$^{1}$ \quad Xiaojuan Qi$^{1}$ \quad Guandao Yang$^{2}$\\[2pt]
{\normalsize $^{1}$The University of Hong Kong \qquad $^{2}$The University of Texas at Austin}
}

\begin{document}
\twocolumn[{%
\renewcommand\twocolumn[1][]{#1}%
\maketitle
\vspace{-1.2em}
\begin{center}
\includegraphics[width=\linewidth]{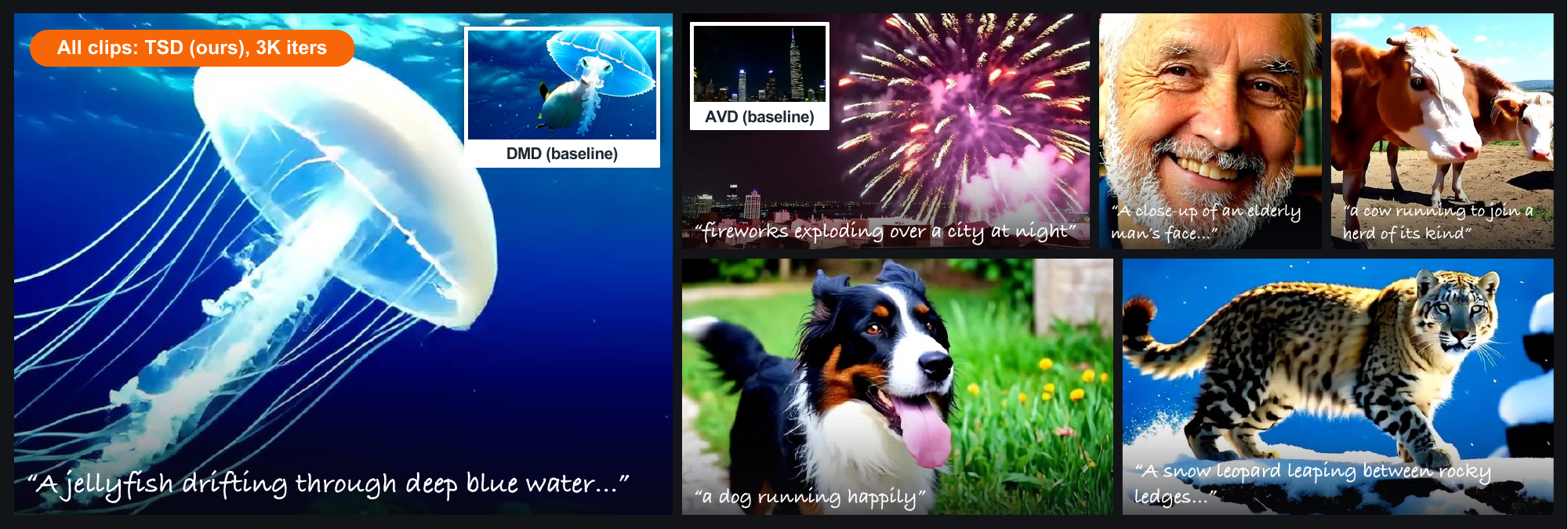}
\vspace{-1.4em}
\captionof{figure}{\textbf{\method{} keeps continued distillation vivid.} Clips from \method{} after 3K iterations, labeled with the opening words of their prompts. Each inset shows a baseline on the same prompt: DMD turns the jellyfish into a fish, and AVD loses the fireworks to a skyline.}
\label{fig:teaser}
\end{center}
\vspace{0.6em}
}]
\begin{abstract}

Reliable video generation is central to interactive applications and digital content creation. Although distribution matching distillation (DMD) enables few-step video generation, balancing visual fidelity with coherent motion remains challenging during prolonged training, and temporal variation may even collapse. We present Temporal Subspace Distillation (TSD), a general supervision framework designed to make better use of teacher knowledge during continued training. Our key idea is to decompose video latents into a temporal mean replicated across frames and zero-mean deviations. These components lie in orthogonal subspaces, enabling separate supervision of persistent visual content and temporal variation. Building on this decomposition, we introduce an identity anchor to transfer time-constant content from the teacher and one-sided deviation floors to penalize insufficient temporal variation in the student. We further extract motion saliency from the same teacher predictions to guide where variation is protected without prescribing a specific motion trajectory. Experiments show that TSD improves video quality over baseline methods, producing videos with high visual fidelity and coherent, plausible motion. TSD retains the student architecture and inference procedure and adds no inference cost. Project page: \url{https://ingra14m.github.io/TSD-website/}.
\end{abstract}

\section{Introduction}
\label{sec:intro}

Video diffusion models synthesize detailed, temporally coherent content, but their iterative sampling is costly for interactive use. Distribution Matching Distillation (DMD) reduces sampling to a few steps by matching a student's distribution to that of a pretrained teacher~\citep{yin2024dmd,yin2024dmd2} and is widely used in autoregressive and real-time video generation~\citep{causvid,selfforcing,causalforcing}. Its training can be unstable~\citep{yin2024dmd2}, and video recipes keep DMD stages short~\citep{selfforcing,causalforcing}. With a 14B Wan2.1 teacher and a 1.3B causal student, longer training sharpens frames while motion degrades into nearly static outputs, temporal discontinuities, or implausible interactions (Figure~\ref{fig:training-motivation}): between 1K and 3K iterations, the dynamic degree of DMD falls from $0.76$ to $0.17$ (Table~\ref{tab:main}).

\begin{figure*}[t]
\centering
\includegraphics[width=\linewidth]{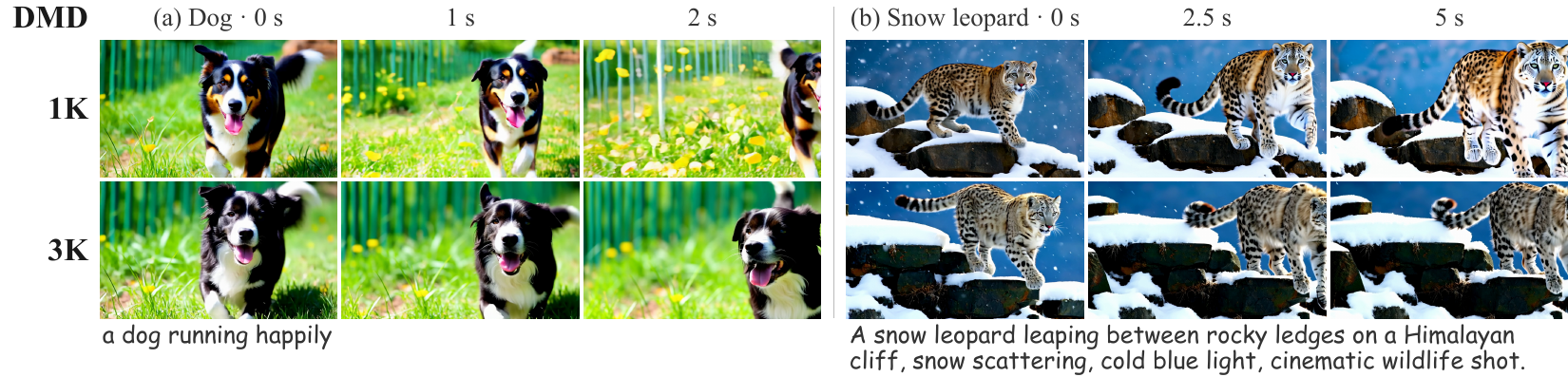}
\caption{\textbf{Continued distillation.} DMD at 1K \vs 3K iterations: the 3K samples show finer details but less apparent, implausible motion.}
\label{fig:training-motivation}
\end{figure*}

Penalizing low temporal variance, as AVD~\citep{avd} does, slows this collapse but costs fidelity, and the teacher prediction that DMD already computes can restore fidelity as an anchor wherever it is a reliable target. Splitting each video latent into its temporal mean, replicated across frames, and the zero-mean deviations from it, two subspaces orthogonal in output space, we measure what the teacher preserves in each (Sec.~\ref{sec:diagnosis}). Its one-step clean prediction preserves the coarse temporal mean at every tested noise level but contracts the deviations increasingly with noise, and its denoising loss is lower on static and time-averaged versions of a clip. These measurements suggest that a teacher target helps in the temporal mean and costs motion in the deviations.

\begin{figure*}[!t]
\centering
\includegraphics[width=0.95\linewidth]{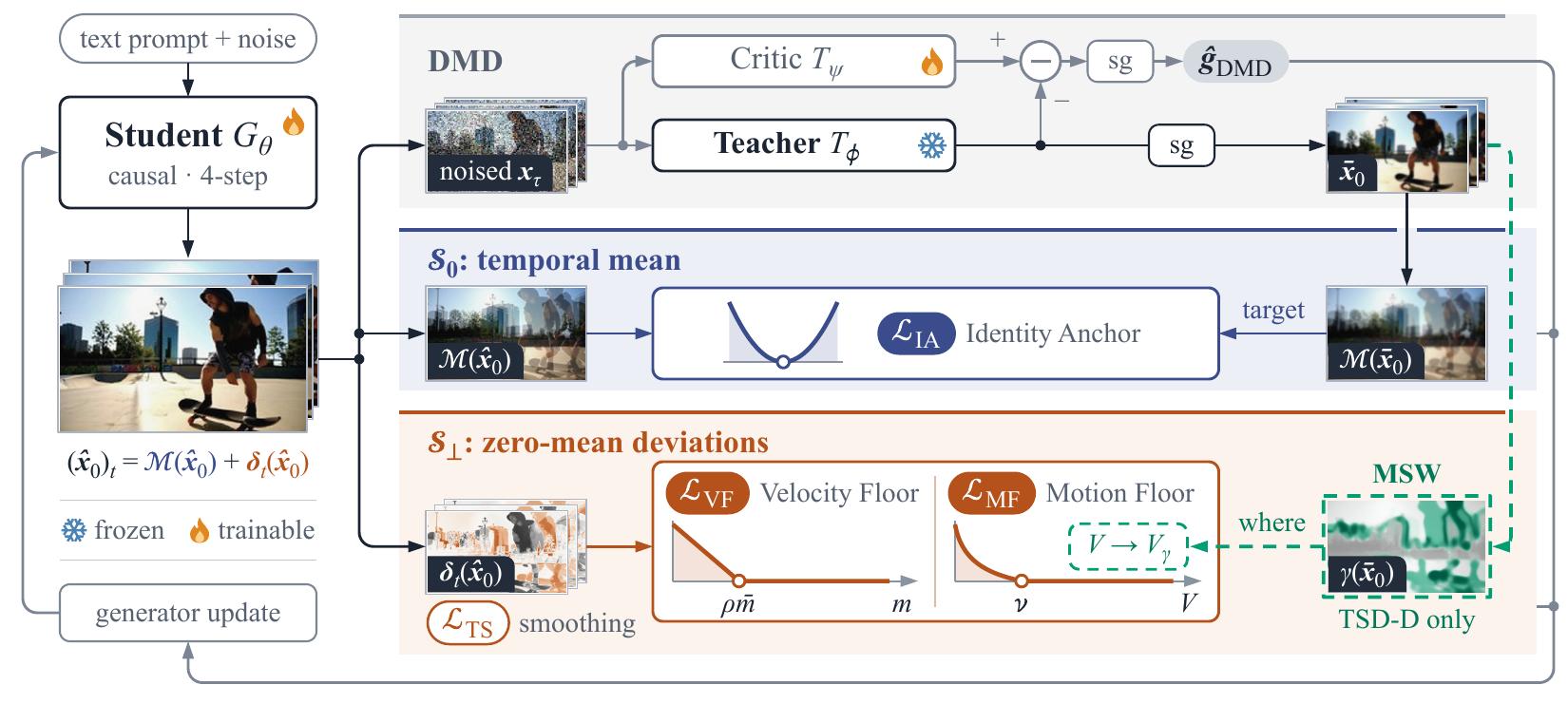}
\caption{\textbf{Overview of \method{}.} The unchanged DMD update acts on both temporal subspaces. The detached teacher prediction DMD already computes supplies the temporal-mean target of the Identity Anchor (IA), the only auxiliary term with a teacher target. In the deviation subspace, Motion Floor (MF) and Velocity Floor (VF) protect the student's own variation against a reference, and Temporal Smoothing (TS) damps flicker. Green: \methodD{}'s saliency weighting (MSW), turning $V$ in MF into $V_\gamma$.}
\label{fig:method}
\end{figure*}

We therefore add teacher supervision only for the temporal mean and protect the deviations with floors on the student's own statistics instead of a teacher target. Temporal Subspace Distillation (\method{}) does so on top of the unchanged DMD update (Figure~\ref{fig:method}): an \emph{Identity Anchor} (IA) ties the student's temporal mean, the time-invariant content that sets a clip's identity, to that of the teacher prediction, adding the critic-free score-distillation direction within the mean subspace (Sec.~\ref{sec:decomp}), and one-sided floors on the student's temporal variance (\emph{Motion Floor}, MF) and low-pass temporal change (\emph{Velocity Floor}, VF) penalize its own variation only below a reference, without prescribing a trajectory. With light \emph{Temporal Smoothing} (TS) they form \methodB{} (Balanced), the simpler variant that our ablations build on. Our default, \methodD{} (Dynamic), weights MF by teacher-derived motion saliency (MSW) and estimates the VF reference from all data-parallel ranks rather than from one clip per rank (CRS).

Over three training seeds, \methodD{} keeps a 3K dynamic degree of $0.85$ (DMD $0.17$, teacher $0.69$) and the highest \vbone{}, and \methodB{} tops \vbtwo{}. Anchoring the full teacher prediction lowers motion, and the floors alone keep motion but lose fidelity. With the same loss weights, \methodD{} transfers to a bidirectional Wan2.1-1.3B student, on par with or above released rCM, AVD, and AnyFlow checkpoints, and to CogVideoX-5B, where DMD collapses likewise. As \vbone{} can reward frozen or shuffled clips, a blind pairwise user study on 64 \vbone{} prompts is our primary evidence: \methodD{}@3K wins 87--95\% of votes against DMD at 3K and at 1K, the checkpoint early stopping would select, and 72--73\% against its own 1K checkpoint, so continued training pays off (Table~\ref{tab:user-study}). 

Overall, our contributions are:
\begin{itemize}[leftmargin=1.4em,itemsep=1pt,topsep=1pt]
\item We propose Temporal Subspace Distillation (TSD), a framework that explicitly decomposes video latents into two orthogonal temporal subspaces for targeted supervision of persistent content and temporal variation.
\item We introduce an identity anchor and a family of one-sided latent deviation floors to improve content learning and mitigate motion degradation during continued distillation.
\item Three-seed benchmarks on two model families with stronger baselines, and a blind user study whose ablations isolate the anchor, VF and TS, and the mean-only restriction.
\end{itemize}

\section{Related Work}
\label{sec:related}

\paragraph{Video generation.}
Video diffusion models~\citep{videodiffusion} moved from U-Nets in pixel or latent space~\citep{imagenvideo,makeavideo,videoldm,svd} to diffusion transformers on latent tokens~\citep{dit,walt,latte,sora}, which underlie most large video models~\citep{opensora,cogvideox,moviegen,hunyuanvideo,stepvideo,wan,seedance,veo3,kling}, many trained with flow matching~\citep{flowmatching,rectifiedflow}. Autoregressive models generate frames or chunks from preceding context, with per-token noise levels~\citep{diffusionforcing}, chunk-wise causal attention~\citep{magi1}, per-frame noise schedules~\citep{skyreelsv2}, or interactive streaming~\citep{longlive,longlive2}.

\paragraph{Diffusion distillation.}
Distillation compresses sampling into a few steps through trajectory approximation, consistency learning, or distribution matching~\citep{progressivedistill,consistencymodels,yin2024dmd}. DMD~\citep{yin2024dmd} matches distributions with a pretrained teacher score and an online estimate of the student's score, plus a paired regression regularizer, which DMD2~\citep{yin2024dmd2} removes while stabilizing training with two-time-scale updates. Other work uses per-phase noise-level experts~\citep{phaseddmd} or adaptive matching objectives~\citep{amd}, or restores diversity and fidelity~\citep{dfd}. These change the matching objective or its schedule, whereas \method{} keeps the DMD update unchanged and adds output losses. rCM~\citep{rcm} regularizes continuous-time consistency distillation with score distillation, and AnyFlow~\citep{anyflow} uses on-policy flow-map distillation. Adversarial distillation~\citep{add,apt} and motion discriminators~\citep{mogan} add a separate network. \method{} adds none. For video, distribution matching yields few-step autoregressive students~\citep{causvid,selfforcing,causalforcing}: CausVid distills a bidirectional teacher into a causal student, Self Forcing trains on self-generated rollouts, and Causal Forcing applies causal ODE distillation followed by DMD with a bidirectional teacher, whose final stage we build on.

\paragraph{Content and motion.}
Separating content from motion has a long history, from decomposed latents~\citep{mocogan} and separate encoders~\citep{mcnet} to accelerated video diffusion, where MCM~\citep{mcm} combines motion distillation with image adversarial learning and AnimateLCM~\citep{animatelcm} decouples image and motion consistency learning. AVD~\citep{avd} pairs adaptive regression with temporal-variance regularization. Other work improves motion with joint appearance-motion targets~\citep{videojam}, inference-time guidance that reduces patch-wise temporal variance~\citep{flowmo}, or reward feedback~\citep{t2vturbo}. Neither MCM nor AVD confines its appearance supervision to the temporal mean, so it can also modify frame-to-frame variation. \method{} keeps AVD's variance penalty, in hinge form, as its motion floor and adds a mean-only teacher anchor (no real videos needed), low-pass velocity protection, temporal smoothing, and saliency weighting (see the supplementary material). Score-distillation methods decompose or mask the gradient along conditioning or spatial axes~\citep{nfsd,semanticsds} rather than time.

\section{Method}
\label{sec:method}

\subsection{Video Distribution Matching}
\label{sec:video-dmd}

DMD matches a generator's distribution to a pretrained teacher through a score difference motivated by reverse-KL minimization~\citep{yin2024dmd,yin2024dmd2}. A few-step causal generator $G_\theta$ maps noise and text (omitted below) to a latent video $\xhat\in\mathbb{R}^{F\times C\times H\times W}$ with $F$ frames, $C$ channels, height $H$, and width $W$. At noise level $\tau$, the noised video is $\vx_\tau=\alpha_\tau\xhat+\sigma_\tau\boldsymbol{\epsilon}$ with Gaussian $\boldsymbol{\epsilon}$. The frozen teacher $\phi$ supplies the guided target score $\sreal$, and an online diffusion critic $\psi$, trained to denoise student samples, estimates the student's score $\sfake$. Both denoise $\vx_\tau$ into clean estimates $T_\phi$ and $T_\psi$. Following Causal Forcing~\citep{causalforcing}, the normalized DMD direction for one clip is
\begin{equation}
\label{eq:dmd-grad}
\widehat{\vg}_{\mathrm{DMD}}
=\frac{T_\psi-T_\phi}{n}
=\frac{\sigma_\tau^2}{\alpha_\tau n}\left[\sfake-\sreal\right],
\end{equation}
where $n=\|\xhat-T_\phi\|_1/(FCHW)$ is the mean absolute discrepancy between the student sample and the teacher estimate (score conversion and guidance in the supplementary material). The direction enters the stop-gradient surrogate~\citep{yin2024dmd,yin2024dmd2}
\begin{equation}
\mathcal{L}_{\mathrm{DMD}}
=\frac{1}{2FCHW}\left\|\xhat-
\sg{\xhat-\widehat{\vg}_{\mathrm{DMD}}}\right\|_2^2,
\end{equation}
with gradient $\widehat{\vg}_{\mathrm{DMD}}/(FCHW)$, where $\sg{\cdot}$ detaches its argument, including the score predictions and $n$, and $\|\cdot\|_2$ is the Frobenius norm. This video-level update supervises persistent content and temporal variation jointly.

\subsection{What the Teacher Preserves}
\label{sec:diagnosis}

\paragraph{Temporal subspaces.}
For a latent video $\vx$ with frames $\vx_t$, $t=1,\dots,F$, the temporal mean is $\Tmean(\vx)=\frac1F\sum_{t}\vx_t$ and the deviations are $\dev_t(\vx)=\vx_t-\Tmean(\vx)$. Broadcasting the mean to all frames gives a time-constant video in the mean subspace $\mathcal S_0$. The deviations $\dev(\vx)$ sum to zero over frames and lie in the orthogonal complement $\mathcal S_\perp$. Let $\Pi_0$ be the mean-broadcasting projection and $\Pi_\perp=I-\Pi_0$. Isotropic noise places only a fraction $1/F$ of its energy in $\mathcal S_0$, so for strongly correlated frames the temporal mean should be the easiest component for one denoising step to recover. We test this on Wan2.1 (Table~\ref{tab:diagnosis}, full tables and protocols in the supplementary material).

\begin{table}[t]
\centering
\setlength{\abovecaptionskip}{0pt}
\setlength{\belowcaptionskip}{4pt}
\caption{\textbf{What the teacher preserves} on Wan2.1. (a)~Gain of the teacher's guided clean prediction along each component of re-noised teacher samples ($1$: preserved, $<1$: contracted, not reconstruction accuracy). $^{\ast}$\methodB{} samples at 3K. (b)~Teacher loss relative to the original clip (median of 24 \methodB{} clips at 3K). (c)~$A_F-A_T$ ($\pm$ 95\% bootstrap half-width) on 72 clips of one DMD run, pooled over noise levels. Positive values mean a DMD step contracts latent temporal variance to first order.}
\label{tab:diagnosis}
{\scriptsize
\setlength{\tabcolsep}{2pt}
\renewcommand{\arraystretch}{1.02}
\begin{tabular*}{\linewidth}{@{\extracolsep{\fill}}lcccc@{}}
\toprule
\textbf{(a)} Teacher gain & $\sigma_\tau{=}0.5$ & $0.9$ & $0.98$ & $0.9^{\ast}$ \\
\midrule
temporal mean, spatial low-pass  & 1.00 & 0.99 & 0.94 & 1.01 \\
temporal mean, spatial high-pass & 0.91 & 0.64 & 0.27 & 0.68 \\
temporal deviations              & 0.91 & 0.72 & 0.45 & 0.83 \\
\midrule
\textbf{(b)} Relative teacher loss & $\sigma_\tau{=}0.2$ & $0.4$ & $0.6$ & $0.8$ \\
\midrule
every frame $\to$ temporal mean & 0.27 & 0.45 & 0.57 & 0.61 \\
every frame $\to$ first frame   & 0.37 & 0.59 & 0.73 & 0.81 \\
spatial blur (2\,px)            & 0.77 & 0.81 & 0.88 & 0.98 \\
\midrule
\textbf{(c)} DMD direction & \multicolumn{3}{c}{$A_F-A_T$ ($\times10^{-3}$)} & clips $>0$ \\
\midrule
1K & \multicolumn{3}{c}{$3.6\pm1.4$} & 0.76 \\
3K & \multicolumn{3}{c}{$1.1\pm1.9$} & 0.50 \\
\bottomrule
\end{tabular*}}
\end{table}

\paragraph{The teacher preserves the coarse temporal mean.}
For a clean latent $\vx_0$, a component map $P$, and the teacher's guided clean prediction $\xbar$ from $\vx_\tau$, the gain $g_P=\langle P(\xbar),P(\vx_0)\rangle/\|P(\vx_0)\|_2^2$ is one if the component is preserved along its own direction and below one if contracted. On re-noised teacher samples (Table~\ref{tab:diagnosis}a), the spatially low-pass temporal mean keeps a gain of $0.94$--$1.00$ at every tested noise level, whereas the deviations contract increasingly with noise, to $0.45$ at $\sigma_\tau=0.98$. The high-pass mean contracts as much or more ($0.27$), so only the coarse mean is preserved. The predicted mean varies across seeds about as much as the clips' own mean (see the supplementary material), and the pattern holds on \methodB{} samples (last column). This contraction of low-SNR components is expected for a posterior-mean denoiser, and the measurement locates which part of a video latent is a reliable target.

\paragraph{The teacher's loss favors static content.}
Replacing every frame of a \methodB{} clip by the temporal mean, or by the first frame, lowers the teacher's flow-matching loss at every tested noise level, by more than an unmatched 2-pixel spatial blur does (Table~\ref{tab:diagnosis}b). The loss thus gives no penalty for lost motion. VideoJAM~\citep{videojam} finds the denoising loss nearly insensitive to frame order, and here the teacher's loss goes further and prefers the static versions.

\paragraph{The DMD direction does not restore the deviations.}
Let $A_T$ and $A_F$ be the alignments of the teacher's and the critic's clean predictions with the student's deviations, $A=\langle\dev(\xhat),\dev(T)\rangle/\|\dev(\xhat)\|_2^2$. Since $\dev(\widehat{\vg}_{\mathrm{DMD}})=(\dev(T_\psi)-\dev(T_\phi))/n$, a small output-space descent step along the DMD direction changes $\|\dev(\xhat)\|_2^2$ in proportion to $-(A_F-A_T)$ to first order. In a vanilla DMD run, both predictions contract the student's deviations ($A_T,A_F\in[0.62,0.94]$), and at 1K the teacher does so slightly more: $A_F-A_T$ is positive on $76\%$ of clips and its pooled interval excludes zero (Table~\ref{tab:diagnosis}c). To first order in output space, the DMD step thus slightly contracts latent temporal variance while the student still moves, and the effect vanishes at 3K, after the collapse.

\paragraph{Principle.}
A teacher target should be reliable and should not act on motion. In $\mathcal S_\perp$ it fails both, and neither the teacher nor the DMD direction restores the deviations, so \method{} protects them with one-sided floors on the student's own statistics. In $\mathcal S_0$ the second condition holds exactly, as a mean-only gradient is the same for every frame (Eq.~(\ref{eq:ia})), and the first holds for the coarse mean at every noise level. The high-pass mean is contracted too, but a mean-only target can soften static detail, not motion, so IA anchors the full temporal mean without a spatial filter or cutoff. Sec.~\ref{sec:ablate} tests whether a target in $\mathcal S_\perp$ costs motion.

\subsection{Mean Anchor}
\label{sec:decomp}

The \emph{Identity Anchoring Loss} (IA) matches the student's temporal mean to that of the detached teacher prediction already computed for DMD, $\xbar=\sg{T_\phi(\vx_\tau,\tau)}$:
\begin{equation}
\label{eq:ia}
\mathcal{L}_{\mathrm{IA}}(\theta)
=\frac{1}{CHW}
\left\|\Tmean(\xhat)-\Tmean(\xbar)\right\|_2^2.
\end{equation}
Every latent frame receives the same output gradient, $\frac{2}{FCHW}\Pi_0(\xhat-\xbar)\in\mathcal S_0$, so IA sets no teacher target for the deviations, which the floors of Sec.~\ref{sec:floor} protect.
IA keeps only the mean part of a framewise regression to the same target: as the deviations sum to zero over frames, the cross term vanishes and
\begin{equation}
\label{eq:anchor-split}
\tfrac{1}{FCHW}\bigl\|\xhat-\xbar\bigr\|_2^2
=\mathcal L_{\mathrm{IA}}
+\tfrac{1}{FCHW}\sum_{t=1}^{F}\bigl\|\dev_t(\xhat)-\dev_t(\xbar)\bigr\|_2^2 .
\end{equation}
A full-prediction anchor therefore adds exactly a deviation term, which regresses the student's deviations toward the contracted teacher prediction. Section~\ref{sec:ablate} compares the mean, deviation, and full anchors.

\paragraph{Relation to score distillation.}
Let $\vg=T_\psi-T_\phi$ be the unnormalized DMD direction. Because $\xbar$ is the detached teacher prediction, $\xhat-\xbar=\xhat-T_\phi$ is the clean-prediction direction of critic-free score distillation~\citep{dreamfusion}. With $N=FCHW$ and $\lambda'=2n\lambda_{\mathrm{IA}}$ (see the supplementary material), the combined output gradient is
\begin{equation}
\label{eq:ia-residual}
nN\nabla_{\xhat}\bigl(\mathcal L_{\mathrm{DMD}}{+}\lambda_{\mathrm{IA}}\mathcal L_{\mathrm{IA}}\bigr)
=\Pi_\perp\vg+\Pi_0\bigl(\vg+\lambda'(\xhat-\xbar)\bigr).
\end{equation}
The DMD direction is unchanged in $\mathcal S_\perp$. In $\mathcal S_0$, IA adds the critic-free direction to it. Applying the critic-free direction in both subspaces removes the critic, which degrades image quality (see the supplementary material).

\begin{table*}[!t]
\centering
\setlength{\abovecaptionskip}{0pt}
\setlength{\belowcaptionskip}{5pt}
\caption{Main comparison on Wan2.1-T2V (14B teacher, 1.3B student), mean$_{\pm\text{sample std}}$ over three training seeds. $\Delta$VB1: \vbone{} change from 1K to 3K. \textsc{init}: shared starting checkpoint (one run). Dynamic degree has no arrow: the 14B teacher reaches $0.69$ (see the supplementary material), so we read it as distance from collapse. \textsc{avd-full} and \textsc{dmd}$+$motion GAN (MoGAN-style~\citep{mogan}) use teacher rollouts (see the supplementary material).}
\label{tab:main}
{\small
\newcommand{\mainscore}[2]{#1_{\pm#2}}
\renewcommand{\arraystretch}{1.0}
\setlength{\tabcolsep}{5pt}
\begin{tabular}{@{}lccccccc@{}}
\toprule
& \multicolumn{2}{c}{\textbf{\vbone{}}\,$\uparrow$} & \multicolumn{2}{c}{\textbf{\vbtwo{}}\,$\uparrow$} & & \multicolumn{2}{c}{\textbf{dyn.\ degree}} \\
\textbf{Method} & 1K & 3K & 1K & 3K & $\Delta$VB1 & 1K & 3K \\
\midrule
\textsc{init} (step 0) & \multicolumn{2}{c}{$84.31$} & \multicolumn{2}{c}{$44.87$} & --- & \multicolumn{2}{c}{$0.82$} \\
\midrule
\textsc{dmd} & $\mainscore{86.43}{0.61}$ & $\mainscore{80.82}{0.62}$ & $\mainscore{50.40}{1.41}$ & $\mainscore{51.98}{1.33}$ & $-5.61$ & $\mainscore{0.76}{0.21}$ & $\mainscore{0.17}{0.09}$ \\
\textsc{avd} & $\mainscore{81.80}{1.86}$ & $\mainscore{81.39}{2.50}$ & $\mainscore{48.17}{2.01}$ & $\mainscore{50.12}{1.44}$ & $-0.41$ & $\mainscore{1.00}{0.00}$ & $\mainscore{1.00}{0.00}$ \\
\textsc{avd-official} & $\mainscore{84.17}{3.67}$ & $\mainscore{82.95}{1.54}$ & $\mainscore{49.73}{3.22}$ & $\mainscore{51.44}{3.33}$ & $-1.22$ & $\mainscore{0.99}{0.01}$ & $\mainscore{0.73}{0.28}$ \\
\textsc{avd-full} & $\mainscore{85.12}{1.40}$ & $\mainscore{83.20}{1.21}$ & $\mainscore{50.21}{1.62}$ & $\mainscore{51.66}{1.31}$ & $-1.92$ & $\mainscore{0.97}{0.02}$ & $\mainscore{0.69}{0.15}$ \\
\textsc{dmd}$+$motion GAN & $\mainscore{86.08}{0.71}$ & $\mainscore{83.41}{1.02}$ & $\mainscore{50.92}{1.20}$ & $\mainscore{51.87}{1.14}$ & $-2.67$ & $\mainscore{0.88}{0.07}$ & $\mainscore{0.58}{0.12}$ \\
\midrule
\methodB{} & $\mainscore{87.00}{0.42}$ & $\mainscore{84.97}{1.13}$ & $\mainscore{\mathbf{52.69}}{1.71}$ & $\mainscore{\mathbf{53.93}}{1.04}$ & $-2.03$ & $\mainscore{0.98}{0.02}$ & $\mainscore{0.71}{0.10}$ \\
\methodD{} & $\mainscore{\mathbf{87.22}}{0.20}$ & $\mainscore{\mathbf{86.86}}{0.48}$ & $\mainscore{52.50}{0.73}$ & $\mainscore{52.05}{1.48}$ & $-0.36$ & $\mainscore{0.99}{0.01}$ & $\mainscore{0.85}{0.05}$ \\
\bottomrule
\end{tabular}%
}
\end{table*}

\subsection{Deviation Protection}
\label{sec:floor}

One-sided floors on the student's own statistics penalize a statistic only below a reference, so they never pull toward a particular trajectory.

\paragraph{Motion Floor (MF).}
With the global temporal variance $V(\vx)=\|\dev(\vx)\|_2^2/(FCHW)$, MF applies a one-sided floor to a variance statistic $V_*$ of the student:
\begin{equation}
\label{eq:floor-global}
\mathcal{L}_{\mathrm{MF}}(V_*)
=\max\!\left\{0,-\log(V_*+\varepsilon)-\tau_{\mathrm v}\right\},
\end{equation}
where $\varepsilon>0$ ensures numerical stability, and the threshold $\tau_{\mathrm v}$, shared across prompts, sets the reference $\nu=e^{-\tau_{\mathrm v}}$: the penalty is active only when $V_*+\varepsilon<\nu$. When active, an output-space descent step on the global floor $V_*=V(\xhat)$ expands existing deviations along their current direction. Otherwise its gradient vanishes.

\paragraph{Saliency weighting (MSW).}
Global MF weights all locations equally, so fluctuations in regions that should stay static can satisfy it. \methodD{} therefore uses the saliency-weighted variance
\begin{equation}
\label{eq:msw-floor}
V_\gamma(\xhat)=\frac{\sum_{t,c,h,w}\gamma_{h,w}\dev_{t,c,h,w}(\xhat)^2}
{(F-1)C\left(\sum_{h,w}\gamma_{h,w}+\varepsilon_\gamma\right)},
\end{equation}
whose detached weights $\gamma_{h,w}\in(0,1)$ are the time-averaged sigmoid of the teacher prediction's normalized frame-to-frame change at $(h,w)$ (see the supplementary material). They indicate where the teacher predicts change, not its direction or timing, so the teacher weights where the floor acts but sets no deviation target.

\paragraph{Velocity Floor (VF).}
Rapid frame-to-frame fluctuations can satisfy the variance reference even when sustained change is weak, so VF protects change that remains after temporal smoothing, across a longer frame interval. Let $\mathcal B$ be a linear temporal blur with normalized Gaussian weights and replicated endpoints, which leaves time-constant signals unchanged. For a stride $1\leq s<F$ and $\mathbf d=\mathcal B\,\dev(\xhat)$, the low-pass change of the deviations is
\begin{equation}
\label{eq:velmag}
m(\xhat)
=\frac{1}{s(F-s)CHW}\sum_{t=1}^{F-s}
\bigl\|\mathbf d_{t+s}-\mathbf d_{t}\bigr\|_1.
\end{equation}
A reference $\bar m$ is estimated from detached student statistics with an exponential moving average (momentum $0.9$), kept per data-parallel rank in \methodB{}, where each rank holds one clip per update at our batch size, and averaged across ranks in \methodD{} (cross-rank synchronization, CRS), and frozen after a warm-up of $T_0$ generator updates so that later contraction cannot lower it (see the supplementary material). VF penalizes change below a fraction $\rho\in(0,1]$ of it:
\begin{equation}
\label{eq:amp}
\mathcal{L}_{\mathrm{VF}}(\theta)=\relu{\rho\bar m-m(\xhat)},
\end{equation}
where $[z]_+=\max(z,0)$.

\paragraph{Temporal Smoothing (TS).}
VF does not penalize the rapid fluctuations removed by its blur. TS penalizes this residual, $\mathcal{L}_{\mathrm{TS}}=\|(I-\mathcal B)\xhat\|_2^2/(FCHW)$, with a small weight to limit its effect on fast coherent motion.

\paragraph{Orthogonality.}
Each deviation term is invariant to time-constant offsets of $\xhat$, so its output gradient lies in $\mathcal S_\perp$, orthogonal to the IA gradient (see the supplementary material). This holds only in output space. DMD acts on both subspaces, and all terms share the generator.

\subsection{Training Objective}
\label{sec:obj}

The generator combines the unchanged DMD loss with mean anchoring and deviation protection,
\begin{equation}
\label{eq:total}
\mathcal{L}(\theta)
=\mathcal{L}_{\mathrm{DMD}}
+\textstyle\sum_{k}\lambda_{k}\mathcal{L}_{k},\quad k\in\{\mathrm{IA},\mathrm{MF},\mathrm{VF},\mathrm{TS}\},
\end{equation}
with $\mathcal{L}_{\mathrm{MF}}$ evaluated at $V_*$. The floors are soft penalties. Both configurations share Eq.~(\ref{eq:total}) and all weights. The simpler \methodB{} (\emph{Balanced}) uses $V_*=V(\xhat)$ and a per-rank VF reference. Our default \methodD{} (\emph{Dynamic}) uses $V_*=V_\gamma(\xhat)$ (MSW) and a rank-shared VF reference (CRS).

\section{Experiments}
\label{sec:exp}

\begin{figure*}[!t]
\centering
\includegraphics[width=0.95\linewidth]{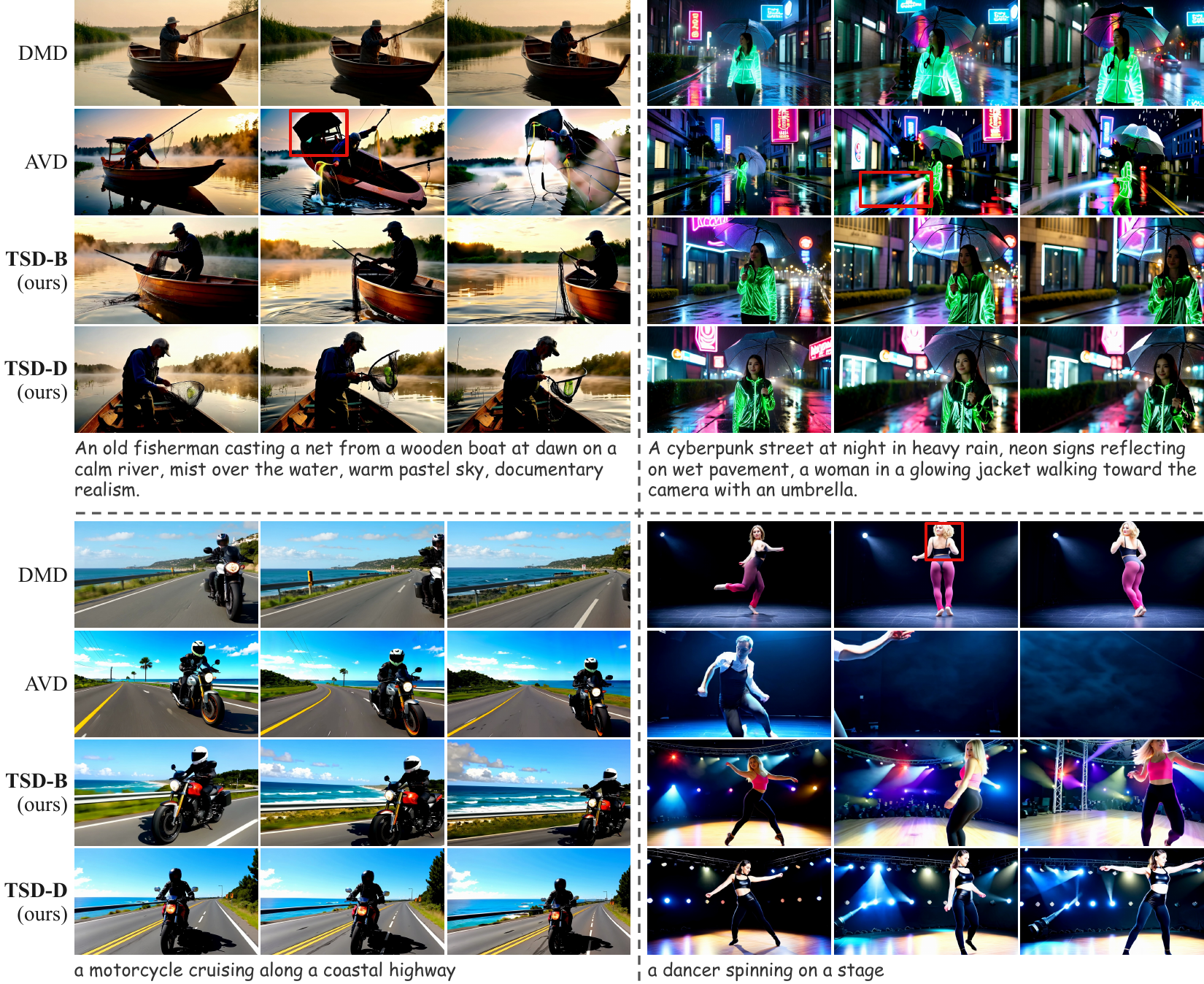}
\caption{Qualitative comparisons. Each method row shows three frames in temporal order, with the prompt below each scene. Upper panels use long scene descriptions and lower panels short action prompts. Red boxes mark distortions and implausible motion.}
\vspace{-5pt}
\label{fig:qual-compare}
\end{figure*}

\subsection{Experimental Setup}
\label{sec:setup}

We distill the frozen bidirectional Wan2.1-T2V-14B~\citep{wan} into a 1.3B causal student initialized from the stage-2 ODE checkpoint of Causal Forcing~\citep{causalforcing}, which generates $81$ frames at $16$\,fps and $832\times480$ with 4 denoising steps per latent frame. All methods share the teacher, initialization, and filtered VidProM~\citep{vidprom} prompts. We train for $3000$ iterations (total batch $8$, eight GPUs), updating the critic every iteration and the student every fifth, so 1K and 3K iterations are 200 and 600 student updates. Both configurations use $\lambda_{\mathrm{IA}}=0.075$, $\lambda_{\mathrm{MF}}=0.05$, $\lambda_{\mathrm{VF}}=0.1$, and $\lambda_{\mathrm{TS}}=0.02$, and \methodD{} adds MSW and CRS. MF uses $\tau_{\mathrm v}=0.8$, and VF uses $\rho=0.9$ with a warm-up of $T_0=200$ generator updates. Every method keeps the guidance, critic ratio, and learning rates of the published stage-3 DMD recipe of Causal Forcing at this batch, so the added losses are compared under one fixed recipe (all other hyperparameters in the supplementary material), and DMD collapses likewise with the teacher and guidance of the second setting (Sec.~\ref{sec:transfer}). We report \vbone{}~\citep{vbench} and \vbtwo{}~\citep{vbench2} with their official prompts and dimension scores, as means over three training seeds unless a single run is stated. Baselines are vanilla DMD, two prompt-only AVD~\citep{avd} variants that keep only its temporal-variance regularizer, always-on (\textsc{avd}) and gated as in the official code (\textsc{avd-official}), and two that use teacher rollouts, AVD with its regression term (\textsc{avd-full}) and a motion discriminator (\textsc{dmd}$+$motion GAN).

\subsection{Human Evaluation}
\label{sec:qualitative}

In Figure~\ref{fig:qual-compare}, vanilla DMD leaves the fisherman almost static and AVD distorts him and the boat, whereas \method{} moves them coherently (1K \vs 3K in the supplementary material).

\begin{table}[t]
\centering
\setlength{\abovecaptionskip}{0pt}
\setlength{\belowcaptionskip}{4pt}
\caption{\textbf{User study} (64 participants, 64 \vbone{} prompts $\times$ 2 seeds). W/T/L: \% of votes for the reference, tie, or the alternative. $^{*}$Holm-corrected $p<0.05$, two-way cluster-robust score test over participants and prompts on non-tie votes. Underlined: pre-specified criterion. Teacher row descriptive.}
\label{tab:user-study}
{\footnotesize
\setlength{\tabcolsep}{2.2pt}
\begin{tabular}{@{}lccccccc@{}}
\toprule
& & \multicolumn{3}{c}{\textbf{Visual quality}} & \multicolumn{3}{c}{\textbf{Motion}} \\
\cmidrule(lr){3-5}\cmidrule(l){6-8}
& $n$ & W & T & L & W & T & L \\
\midrule
\multicolumn{8}{@{}l}{\textbf{\methodD{}@3K vs.}} \\
\textsc{dmd}@1K            & 128 & 89.1$^{*}$ & 3.1  & 7.8  & 89.8$^{*}$ & 6.3  & 3.9  \\
\textsc{dmd}@3K            & 128 & 87.5$^{*}$ & 0.8  & 11.7 & 95.3$^{*}$ & 0.0  & 4.7  \\
\textsc{avd-official}@3K   & 128 & 97.7$^{*}$ & 0.8  & 1.6  & 98.4$^{*}$ & 0.0  & 1.6  \\
\textsc{dmd}+motion GAN@3K & 128 & 85.9$^{*}$ & 1.6  & 12.5 & 86.7$^{*}$ & 0.8  & 12.5 \\
\methodD{}@1K              & 128 & 73.4$^{*}$ & 0.0  & 26.6 & 71.9$^{*}$ & 3.9  & 24.2 \\
14B teacher                & 128 & 39.1       & 13.3 & 47.7 & 44.5       & 15.6 & 39.8 \\
\midrule
\multicolumn{8}{@{}l}{\textbf{\methodB{}@3K vs.}} \\
w/o IA                     & 384 & \underline{75.8}$^{*}$ & 2.9 & 21.4 & 76.3 & 2.6 & 21.1 \\
w/o VF, TS                 & 384 & 73.4 & 3.6  & 22.9 & \underline{77.3}$^{*}$ & 2.3  & 20.3 \\
Full anchor                & 384 & 35.7 & 30.2 & 34.1 & \underline{51.8}$^{*}$ & 28.1 & 20.1 \\
\bottomrule
\end{tabular}}
\end{table}

\paragraph{User study.}
Each of 64 participants completed one of 64 questionnaire versions with 30 side-by-side pairs and three attention checks. The pairs come from 64 random \vbone{} prompts, half from its dynamic-degree list and half from its quality lists, each with two noise seeds, and both videos of a pair share the prompt, noise seed, and training seed. For each pair, participants judge which video has better visual quality and which has more natural motion, with a tie option, and motion that is too weak or too strong counts against a video. All responses passed the checks (see the supplementary material). \methodD{}@3K wins 86--98\% of votes on both criteria against every baseline in the study (Table~\ref{tab:user-study}), including DMD at its best checkpoint (1K) and visual quality against DMD@3K, which frame-level \vbone{} dimensions favor. It is also preferred over its own 1K checkpoint despite a lower dynamic degree. Against the 50-step 14B teacher, it receives 45\% and 53\% of the non-tie votes (95\% lower bounds 34\% and 42\%). Every tested preference holds in both the motion and the quality prompt halves (69--100\% of non-tie votes).

\subsection{Benchmark Scores and Transfer}
\label{sec:main}
\label{sec:transfer}

\vbone{} was not designed to rank motion: freezing 14B teacher clips raises its frame-similarity dimensions, and shuffling their frames raises dynamic degree from $0.69$ to $0.97$ (see the supplementary material), so the user study comes first. In Table~\ref{tab:main}, \methodD{} has the highest mean \vbone{} at 1K and 3K, and at 3K every \methodD{} run exceeds every DMD run (see the supplementary material). From 1K to 3K, DMD's dynamic degree collapses and its \vbone{} falls by $5.61$, whereas \methodD{} keeps most of its motion and \vbone{}, which gives it its lead over DMD at 3K. Against DMD's best checkpoint (1K), \methodD{}@3K is on par on \vbone{} and about one standard deviation higher on \vbtwo{}. \textsc{avd-full} and the motion discriminator slow the collapse but trail \methodD{} at 3K on \vbone{} and dynamic degree, and the discriminator adds a trainable network.

\paragraph{Second setting.}
With the same loss weights, we distill bidirectional Wan2.1-T2V-1.3B from itself to four NFE (AnyFlow protocol, see the supplementary material). DMD peaks at 1.5K ($82.70$) and degrades by 5K ($76.90$, dynamic degree $0.84\to0.41$), whereas \methodD{} at 5K keeps a dynamic degree of $0.91$ and reaches $83.62$, above the teacher ($83.06$), the best DMD checkpoint and the released rCM ($83.13$) and AVD ($81.47$, with its regression term) checkpoints, and on par with AnyFlow ($83.54$).

\begin{table}[t]
\centering
\setlength{\abovecaptionskip}{0pt}
\setlength{\belowcaptionskip}{4pt}
\caption{\textbf{Second model family}: CogVideoX-5B~\citep{cogvideox} distilled from itself into a bidirectional 4-step student (49 frames, $480\times720$, three training seeds, main-setting hyperparameters with floor references re-estimated from the initialization). Flow in pixels.}
\label{tab:third}
{\footnotesize
\newcommand{\cs}[2]{\ensuremath{#1_{\pm#2}}}
\setlength{\tabcolsep}{1pt}
\begin{tabular*}{\linewidth}{@{\extracolsep{\fill}}lccc@{}}
\toprule
\textbf{Method} & \textbf{dyn.} & \textbf{Flow} & \textbf{\vbone{}} \\
\midrule
\textsc{dmd}, 1K & \cs{0.38}{0.05} & $5.1$ & $81.82$ \\
\textsc{dmd}, 3K & \cs{0.21}{0.08} & $3.6$ & $78.97$ \\
\textsc{avd-official}, 3K & \cs{0.69}{0.18} & $7.4$ & $80.47$ \\
\methodD{}, 3K & \cs{0.77}{0.10} & $8.6$ & $\mathbf{82.95}$ \\
\bottomrule
\end{tabular*}}
\end{table}

\paragraph{Second model family.}
To test transfer beyond Wan2.1, we distill CogVideoX-5B~\citep{cogvideox}, with a different transformer, VAE, and frame count, into a bidirectional 4-step student (Table~\ref{tab:third}). DMD again loses motion between 1K and 3K, whereas \methodD{} at 3K keeps a dynamic degree of $0.77$ and the highest \vbone{}, and in a separate blind study (20 participants, 24 prompts with two noise seeds each) participants prefer it over DMD at 1K on visual quality ($90.4\%$ of non-tie votes) and motion ($86.9\%$, both $p<0.01$).

\subsection{Ablation}
\label{sec:ablate}

\begin{table}[t]
\centering
\setlength{\abovecaptionskip}{0pt}
\setlength{\belowcaptionskip}{4pt}
\caption{\textbf{Teacher-anchor ablation} (ODE initialization, three training seeds, 3K, detached target $\xbar$, and $\lambda=0.075$ unless noted). The full anchor is the sum of the mean and deviation anchors (Eq.~(\ref{eq:anchor-split})). Rows with floors use those of \methodB{} (no MSW or CRS). w/o dyn.: the total with dynamic degree removed.}
\label{tab:anchor}
{\footnotesize
\newcommand{\cs}[2]{\ensuremath{#1_{\pm#2}}}
\setlength{\tabcolsep}{1pt}
\begin{tabular*}{\linewidth}{@{\extracolsep{\fill}}lcccc@{}}
\toprule
\textbf{Anchor} & \textbf{dyn.\ degree} & \textbf{\vbone{}} & \textbf{w/o dyn.} & \textbf{\vbtwo{}} \\
\midrule
\multicolumn{5}{@{}l}{\emph{Without floors (vanilla \textsc{dmd} $+$ anchor)}} \\
none & \cs{0.17}{0.09} & \cs{80.82}{0.62} & \cs{84.16}{0.52} & \cs{51.98}{1.33} \\
mean & \cs{0.27}{0.18} & \cs{81.88}{1.07} & \cs{83.90}{1.85} & \cs{49.33}{1.95} \\
full & \cs{0.23}{0.14} & \cs{82.60}{0.51} & \cs{85.09}{1.26} & \cs{50.11}{1.58} \\
\midrule
\multicolumn{5}{@{}l}{\emph{With MF $+$ VF $+$ TS (mean row $=$ \methodB{})}} \\
none & \cs{0.74}{0.14} & \cs{82.77}{1.31} & \cs{78.84}{0.53} & \cs{52.88}{0.82} \\
deviation & \cs{0.53}{0.17} & \cs{81.07}{0.98} & \cs{79.73}{1.20} & \cs{51.76}{1.35} \\
full, $\lambda/4$ & \cs{0.59}{0.07} & \cs{83.02}{0.66} & \cs{80.94}{0.35} & \cs{52.81}{0.79} \\
full & \cs{0.47}{0.42} & \cs{81.83}{4.37} & \cs{81.26}{0.88} & \cs{52.21}{0.34} \\
mean & \cs{0.71}{0.10} & \cs{84.97}{1.13} & \cs{81.48}{0.13} & \cs{53.93}{1.04} \\
\bottomrule
\end{tabular*}}
\end{table}

\begin{table}[t]
\centering
\setlength{\abovecaptionskip}{0pt}
\setlength{\belowcaptionskip}{5pt}
\newcommand{\compactscore}[2]{\ensuremath{#1_{\pm#2}}}
\caption{\textbf{Component ablation} at 3K, three training seeds, ODE initialization. The lower block adds saliency weighting (MSW) and the rank-shared VF reference (CRS) to \methodB{}, giving \methodD{}.}
\label{tab:ablate}
\footnotesize
\renewcommand{\arraystretch}{1.13}
\setlength{\tabcolsep}{0pt}
\begin{tabular*}{\linewidth}{@{\extracolsep{\fill}}lcc@{}}
\toprule
\textbf{Configuration} & \textbf{VB1}\,$\uparrow$ & \textbf{VB2}\,$\uparrow$ \\
\midrule
\textsc{dmd} & \compactscore{80.82}{0.62} & \compactscore{51.98}{1.33} \\
$+$IA & \compactscore{81.88}{1.07} & \compactscore{49.33}{1.95} \\
$+$MF & \compactscore{83.81}{3.69} & \compactscore{52.80}{0.6} \\
$+$IA$+$MF & \compactscore{85.65}{1.16} & \compactscore{52.28}{1.2} \\
$+$IA$+$MF$+$VF & \compactscore{83.64}{2.36} & \compactscore{51.78}{1.2} \\
$+$IA$+$MF$+$TS & \compactscore{81.70}{4.65} & \compactscore{50.64}{5.0} \\
$+$IA$+$MF$+$VF$+$TS $=$ \methodB{} & \compactscore{84.97}{1.13} & \compactscore{\mathbf{53.93}}{1.04} \\
\midrule
\multicolumn{3}{@{}l}{\emph{added to} \methodB{}:} \\
$+$MSW & \compactscore{83.71}{2.50} & \compactscore{52.16}{2.2} \\
$+$CRS & \compactscore{85.89}{0.15} & \compactscore{51.60}{0.7} \\
$+$MSW$+$CRS $=$ \methodD{} & \compactscore{\mathbf{86.86}}{0.48} & \compactscore{52.05}{1.48} \\
\bottomrule
\end{tabular*}
\end{table}

\paragraph{Components.}
Built component by component (Table~\ref{tab:ablate}), the floors carry the motion and the anchor the fidelity. $+$MF alone matches \textsc{avd-official} on \vbone{} within noise, adding IA to MF raises \vbone{}, and VF with TS completes \methodB{}, which has the highest \vbtwo{}. Adding MSW and CRS together gives \methodD{} the highest \vbone{}. Participants confirm IA on visual quality and VF$+$TS on motion, their pre-specified criteria, and \methodB{} wins 73--77\% of votes on either criterion in both comparisons (Table~\ref{tab:user-study}).

\paragraph{Teacher anchor.}
Table~\ref{tab:anchor} changes only the anchor within each block (same weight and seeds). By Eq.~(\ref{eq:anchor-split}), the full and mean rows with floors differ only by the deviation term. Adding it lowers the mean 3K dynamic degree and makes it vary widely across seeds (see the supplementary material), with no frame-level gain and a lower \vbtwo{}, and neither weight of the full anchor ($\lambda$, $\lambda/4$) matches the mean anchor on both. The deviation anchor alone also lowers dynamic degree. Removing the anchor keeps motion but lowers both the frame-level total and \vbtwo{}. Without floors, neither anchor prevents the collapse, and with motion already lost, the full anchor scores slightly higher on the benchmarks. Participants confirm the motion cost: against the full anchor, \methodB{} wins 52\% and loses 20\% of motion votes, while visual quality is even (36\% \vs 34\%, Table~\ref{tab:user-study}). Consistent with Sec.~\ref{sec:diagnosis}, the teacher target helps in the temporal mean and costs motion in the deviations.

\section{Conclusion}
\label{sec:conclusion}

The teacher's clean prediction preserves the coarse temporal mean of a video latent but contracts its deviations, so it suggests a teacher target only for the mean. \method{} anchors only the temporal mean to the teacher and protects the deviations with one-sided floors on the student's own statistics, at no inference cost. \methodD{} keeps a 3K dynamic degree of $0.85$ (DMD $0.17$), and participants prefer it at 3K over DMD at both checkpoints and over \textsc{avd-official} on visual quality and motion.

\paragraph{Limitations.} Our measurements locate where the teacher is a reliable target but do not identify why DMD collapses, and the floors, like AVD's regularizer, address its symptom. The causal runs follow the short DMD stage of current recipes (600 student updates), over which \methodD{}'s dynamic degree falls by $0.14$ and DMD's by $0.59$, and the bidirectional runs extend to 5K iterations. All runs use a total batch of $8$, except for a single-run check at batch $64$ (Appendix~\ref{app:batch64}). Training instability is inherited from DMD. MSW and CRS raise \vbone{}, but our pilot user study detects no perceptual difference between \methodD{} and \methodB{} (Appendix~\ref{app:user-study}). A spatially low-pass mean anchor, which the measurements also suggest, is not evaluated. The floors penalize too little latent variation, not incorrect motion, so camera-like motion can partly satisfy them. Their references are shared across prompts, so a nearly static prompt receives the same floor. Latent variation need not appear as visible motion.

{
    \small
    \bibliographystyle{ieeenat_fullname}
    \bibliography{main}
}

\appendix
\clearpage
\setcounter{page}{1}
\renewcommand{\thepage}{S\arabic{page}}
\maketitlesupplementary
\raggedbottom

\section{Training Details}
\label{app:impl}
\paragraph{Backbone and distillation.} We follow the stage-3 DMD recipe of Causal Forcing~\citep{causalforcing} on Wan2.1-T2V~\citep{wan}. The frozen teacher is the bidirectional 14B model, while the causal student and online critic are both 1.3B. The student generates one frame per causal block in four denoising steps $\{1000,750,500,250\}$ with timestep shift $5.0$. The latent shape is $F\times C\times H\times W=21\times16\times60\times104$, in frame, channel, height, and width order, corresponding to approximately 81 decoded frames at $480\times832$ resolution.

\paragraph{DMD scores and guidance.} The teacher and critic scores in Eq.~(1) of the main paper are $\sreal=(\alpha_\tau T_\phi-\vx_\tau)/\sigma_\tau^2$ and $\sfake=(\alpha_\tau T_\psi-\vx_\tau)/\sigma_\tau^2$. The teacher uses classifier-free guidance $T_\phi=T_c+g(T_c-T_{\mathrm{neg}})$ with $g=3$, where $T_c$ and $T_{\mathrm{neg}}$ are clean predictions conditioned on the input and negative prompts, respectively. The critic uses the input prompt without guidance. Both scores and the normalizer $n$ are detached during the generator update.

\paragraph{Optimization.} We train with AdamW using $(\beta_1,\beta_2)=(0,0.999)$. The generator and critic learning rates are $2\times10^{-6}$ and $4\times10^{-7}$. The generator EMA begins at step $200$ with weight $0.99$.

\paragraph{Objective configuration.} Both configurations, \methodB{} and \methodD{}, use $\lambda_{\mathrm{IA}}=0.075$, $\lambda_{\mathrm{MF}}=0.05$, $\lambda_{\mathrm{VF}}=0.1$, and $\lambda_{\mathrm{TS}}=0.02$. MF uses $\tau_{\mathrm v}=0.8$ and $\varepsilon=10^{-6}$. VF uses $\rho=0.9$, a warm-up of $T_0=200$ generator updates, stride $s=4$, and a temporal Gaussian kernel of size $5$ with width $\sigma_{\mathcal B}=1.5$ frames. \methodD{} additionally enables MSW with $\kappa=3.0$ and $b=1.1$, together with CRS. \methodB{} uses neither. The prompt-only AVD~\citep{avd} baselines use $-\log(V+\varepsilon)$ with weight $0.05$. \textsc{avd-official} gates this term at $\tau_{\mathrm v}=0.8$, and neither variant includes real-video regression.

\paragraph{Relation to AVD.}
Global MF builds on AVD's gated temporal-variance regularizer~\citep{avd} and replaces its gate with a continuous hinge. Let $a=-\log(V+\varepsilon)$. The official AVD implementation applies $a$ when $a\geq\tau_{\mathrm v}$ and zero otherwise, whereas our formulation uses $\max(0,a-\tau_{\mathrm v})$. The hinge approaches zero continuously as the variance reaches its reference, while the gated loss drops discontinuously to zero once the reference is exceeded. \method{} combines this floor with a teacher target restricted to the temporal mean (Secs.~3.2 and~3.3 of the main paper), added on top of the unchanged DMD update, and with low-pass deviation protection, temporal smoothing, and, in \methodD{}, teacher-derived spatial weighting and a rank-shared VF reference. \method{} needs no real videos, whereas AVD's adaptive regression requires them and is omitted in our prompt-only adaptations (Sec.~4.1 of the main paper).

\paragraph{VF warm-up reference and cross-rank synchronization (CRS).} The VF loss in Eq.~(9) of the main paper uses a reference $\bar m$ estimated during warm-up and frozen afterward. In the main setting, 1K iterations are 200 generator updates, so the warm-up of $T_0=200$ ends at the 1K checkpoint. 1K models therefore use only the warm-up reference, and the frozen reference acts only after 1K. CRS changes how this reference is estimated and adds no loss. With one clip per rank (total batch $8$, eight GPUs), each rank's reference in \methodB{} is updated from a single clip per update. Adding CRS to \methodB{} lowers the \vbone{} seed std from $1.13$ to $0.15$ but also lowers \vbtwo{} from $53.93$ to $51.60$ (Table~6 of the main paper). The lower spread is consistent with a less noisy reference, although three seeds do not establish this. Let $m_k^{(r)}$ be the low-pass temporal change statistic in Eq.~(8) of the main paper at generator update $k$ on rank $r$. In \methodD{}, CRS averages the detached statistics across $R$ ranks to update the rank-shared reference:
\begin{equation}
\label{eq:sync}
\begin{aligned}
m_{\mathrm{pool},k}
&=\frac{1}{R}\sum_{r=1}^{R}\sg{m_k^{(r)}},\\
\bar m_1&=m_{\mathrm{pool},1},\\
\bar m_k
&=\beta\bar m_{k-1}+(1-\beta)m_{\mathrm{pool},k},
\qquad 2\leq k\leq T_0,\\
\bar m_k&=\bar m_{T_0},
\qquad k>T_0.
\end{aligned}
\end{equation}
Here $m_{\mathrm{pool},k}$ is the cross-rank average, $\bar m_k$ is the reference at generator update $k$, $\sg{\cdot}$ denotes stop-gradient, and $\beta=0.9$ is the EMA coefficient. \methodB{} instead applies the same recursion independently on each rank using $\sg{m_k^{(r)}}$, so each rank keeps its own reference. The VF loss uses the current $\bar m_k$, including during warm-up.

\paragraph{Saliency weights.} MSW (Sec.~3.4 of the main paper) uses temporal differences in the detached teacher prediction $\xbar$ as a soft spatial mask. For spatial location $(h,w)$ and frame $t\geq2$, let $d_{t,h,w}$ be the channel-averaged teacher change, with $d_{1,h,w}=d_{2,h,w}$. The weight $\gamma_{h,w}$ in Eq.~(7) of the main paper is
\begin{equation}
\label{eq:gate}
\begin{gathered}
d_{t,h,w}(\xbar)
=\frac1C\left\|
(\xbar)_{t,:,h,w}-(\xbar)_{t-1,:,h,w}
\right\|_1,\\
\hat d
=\frac{1}{FHW}\sum_{t,h,w}d_{t,h,w}(\xbar),\\
\gamma_{h,w}
=\frac1F\sum_{t=1}^{F}
\operatorname{sigmoid}\!\left(
\kappa\left[
\frac{d_{t,h,w}(\xbar)}{\hat d+\varepsilon_d}-b
\right]\right).
\end{gathered}
\end{equation}
The colon indexes all channels, $\hat d$ is the mean change over frames and locations, and the sigmoid maps normalized change to weights between zero and one. The scalars $\kappa>0$ and $b$ control weight sharpness and selectivity, and $\varepsilon_d>0$ stabilizes the normalization. In Eq.~(7) of the main paper, $\dev_{t,c,h,w}(\xhat)$ denotes an entry of the deviation tensor and $\varepsilon_\gamma>0$ stabilizes the weighted normalization. Averaging the weights over time indicates where variation is protected without prescribing its direction or timing.

\paragraph{Numerical details.} The MSW variance in Eq.~(7) of the main paper uses the sample-variance normalization $F-1$. Uniform weights recover $\tfrac{F}{F-1}V(\xhat)$ up to the denominator stabilizer, so at the same $\tau_{\mathrm v}$ the weighted floor activates at a global variance about $5\%$ lower for $F=21$. Static teacher predictions produce uniform weights and do not disable MF. We set $\varepsilon_\gamma=\varepsilon/C=10^{-6}/C$ and $\varepsilon_d=10^{-6}$. The first frame difference is copied from the second, giving $d_{1,h,w}=d_{2,h,w}$ at each spatial location $(h,w)$. Temporal Gaussian blurs use replicate padding and leave clips shorter than the kernel unchanged.

\section{Additional Experiments}

\subsection{VBench Sanity Check and Benchmark Scores}
\label{app:vbench}
\begin{table}[t]
\centering
\setlength{\abovecaptionskip}{0pt}
\setlength{\belowcaptionskip}{4pt}
\caption{\textbf{VBench sanity check} on 240 clips of the 50-step 14B teacher (\vbone{} dynamic-degree prompts). Frozen: every frame replaced by the first frame. Shuffled: frame order randomly permuted. Raw \vbone{} dimensions ($0$ to $1$).}
\label{tab:sanity}
{\footnotesize
\setlength{\tabcolsep}{1.5pt}
\begin{tabular*}{\linewidth}{@{\extracolsep{\fill}}lccccccc@{}}
\toprule
\textbf{Clips} & \textbf{subj.} & \textbf{bg.} & \textbf{flick.} & \textbf{smooth.} & \textbf{aesth.} & \textbf{imag.} & \textbf{dyn.} \\
\midrule
Original & $0.954$ & $0.964$ & $0.970$ & $0.983$ & $0.604$ & $0.686$ & $0.69$ \\
Frozen   & $1.000$ & $1.000$ & $1.000$ & $0.998$ & $0.609$ & $0.687$ & $0.00$ \\
Shuffled & $0.935$ & $0.950$ & $0.908$ & $0.915$ & $0.600$ & $0.686$ & $0.97$ \\
\bottomrule
\end{tabular*}}
\end{table}

\vbone{} was not designed to rank motion quality. As a sanity check (Table~\ref{tab:sanity}), freezing each 14B teacher clip at its first frame raises all four frame-similarity dimensions above those of the original clip and leaves aesthetic and imaging quality essentially unchanged, while shuffling the frames destroys coherent motion yet raises dynamic degree from $0.69$ to $0.97$. Frame-level \vbone{} dimensions therefore reward static output, and dynamic degree rewards any change. We report \vbone{} and \vbtwo{} for completeness and comparability, and rely on blind human preference (Sec.~4.2 of the main paper) as primary evidence.

\paragraph{Reading the \vbone{} scores.}
The \vbone{} total includes dynamic degree, and four of its frame-level quality dimensions reward similarity between frames. At 3K, DMD's low-motion clips score higher on these frame-level dimensions, as static clips do in Table~\ref{tab:sanity}, and the 3K lead of \methodD{} over DMD in Table~2 of the main paper comes from its retained dynamic degree. \methodD{} at 3K also matches DMD's best \vbone{} checkpoint (1K) on \vbone{} ($86.86$ \vs $86.43$) and exceeds it on \vbtwo{} ($52.05$ \vs $50.40$). The user study (Sec.~4.2 of the main paper) measures perceived quality and motion directly and is our primary evidence.

\subsection{Component Ablation}
\label{app:ablate-terms}
Table~6 of the main paper builds \method{} component by component over three seeds. $+$MF and \textsc{avd-official} apply the same update when active, so their gap ($83.81$ \vs $82.95$ \vbone{}) reflects run-to-run noise. The mean anchor alone raises \vbone{} but not \vbtwo{}. With MF, it raises \vbone{} to $85.65$. Adding VF or TS alone to IA$+$MF lowers both means, whereas both together (\methodB{}) give the highest \vbtwo{} ($53.93$), so VF and TS are supported only jointly. On top of \methodB{}, MSW alone lowers both scores, CRS alone raises \vbone{} with the smallest spread but lowers \vbtwo{}, and both together give \methodD{}, the highest \vbone{} ($86.86$), at a lower \vbtwo{} than \methodB{}. Relative to CRS alone, adding MSW raises \vbtwo{} only within one standard deviation ($51.60\to52.05$). Several of these differences are within one standard deviation. As Table~6 has no motion column, the component roles stated in Sec.~4.4 of the main paper rest on Table~5 of the main paper: the floors alone reach a dynamic degree of $0.74$, and the mean anchor raises the total without dynamic degree from $78.84$ to $81.48$ but brings no frame-level gain without floors. In the user study, \methodB{} receives most votes on both criteria against the ablations without IA and without VF$+$TS, but only one pre-specified criterion was tested per ablation. The study therefore shows that both are needed but does not separate their roles.

\subsection{Teacher-Anchor Variants}
\label{app:anchor}
The deviation anchor in Table~5 of the main paper is the second term on the right of Eq.~(4) of the main paper, and the full anchor is its left-hand side. Both use the same detached teacher target and weight $\lambda=0.075$ as IA, with the normalizations of Eq.~(4) of the main paper, so the gradient of the full anchor is the sum of the IA and deviation-anchor gradients. All other settings follow \methodB{} (or vanilla DMD for the rows without floors), and every variant uses the same three seeds (s0/r1/r2) as Table~\ref{tab:perseed}. Per-seed 3K dynamic degrees are $0.81$/$0.70$/$0.62$ for the mean anchor, and the full anchor varies most across seeds (Table~5 of the main paper).

\subsection{Second Setting}
\label{app:second}
The second setting distills Wan2.1-T2V-1.3B into a bidirectional 4-NFE student initialized from the same weights, with the 1.3B model as teacher (guidance $5$), 42K VidProM captions, eight GPUs and a total batch of $8$. DMD and \methodD{} share every training hyperparameter of the main setting. The floor references are re-estimated from the initialization in the same way. Evaluation follows AnyFlow~\citep{anyflow}: its 944 augmented \vbone{} prompts with five samples each and its per-sample seeds. The released rCM~\citep{rcm}, AVD~\citep{avd} and AnyFlow checkpoints are evaluated once with their official four-step samplers. Trained rows report the mean$\pm$sample std over three training seeds at the stated checkpoint. The best DMD checkpoint is selected on held-out prompts.

\begin{table}[t]
\centering
\setlength{\abovecaptionskip}{0pt}
\setlength{\belowcaptionskip}{4pt}
\caption{\textbf{Second setting}: bidirectional Wan2.1-T2V-1.3B distilled from itself to 4 NFE (teacher guidance 5), evaluated with the AnyFlow protocol~\citep{anyflow} (944 augmented \vbone{} prompts, five samples each) and the official \vbone{} weights. Trained rows: three training seeds. \methodD{} uses the hyperparameters of the main setting unchanged. Released checkpoints of rCM~\citep{rcm}, AVD~\citep{avd} (with its regression term) and AnyFlow are evaluated once. Dyn.: dynamic degree ($0$ to $1$).}
\label{tab:second}
{\footnotesize
\newcommand{\cs}[2]{\ensuremath{#1_{\pm#2}}}
\setlength{\tabcolsep}{1pt}
\begin{tabular*}{\linewidth}{@{\extracolsep{\fill}}lcccc@{}}
\toprule
\textbf{Method} & \textbf{Total} & \textbf{Quality} & \textbf{Semantic} & \textbf{Dyn.} \\
\midrule
Teacher, $50{\times}2$ NFE & $83.06$ & $85.02$ & $75.23$ & $0.69$ \\
rCM & $83.13$ & $85.14$ & $75.10$ & $0.78$ \\
AVD & $81.47$ & $83.61$ & $72.91$ & $0.64$ \\
AnyFlow & $83.54$ & $85.28$ & $\mathbf{76.57}$ & $0.59$ \\
\midrule
\textsc{dmd}, best (1.5K) & \cs{82.70}{0.31} & $84.98$ & $73.58$ & \cs{0.84}{0.03} \\
\textsc{dmd}, 5K & \cs{76.90}{2.40} & $77.60$ & $74.10$ & \cs{0.41}{0.10} \\
\methodD{}, 5K & \cs{\mathbf{83.62}}{0.28} & $\mathbf{85.93}$ & $74.38$ & \cs{0.91}{0.02} \\
\bottomrule
\end{tabular*}}
\end{table}

\paragraph{Results.}
This setting tests transfer beyond the causal student (Table~\ref{tab:second}, official \vbone{} weights, main-setting loss weights, re-estimated floor references). DMD peaks at 1.5K ($82.70$) and degrades by 5K ($76.90$), its dynamic degree falling from $0.84$ to $0.41$. At 5K, \methodD{} keeps a dynamic degree of $0.91$ and reaches $83.62$, above the best DMD and released AVD ($81.47$, with its regression term) checkpoints, numerically above the teacher ($83.06$) and rCM ($83.13$), and on par with AnyFlow ($83.54$). Its Quality lead over these three is smaller than the contribution of its higher dynamic degree under the official weights ($0.91$ \vs $0.59$--$0.78$), and its Semantic score is the lowest of the four.

\subsection{Additional Baselines and Metrics}
\label{app:baselines}
\textsc{avd-full} keeps AVD's adaptive regression term. Since our training uses text prompts only, its regression targets are 50-step teacher rollouts for 8K training prompts, generated once before training. \textsc{dmd}$+$motion GAN adds a discriminator head on latent frame differences of the critic features, following MoGAN~\citep{mogan}, with the same teacher rollouts as positives and the DMD2 weighting of the adversarial term. Both use the student and critic schedule of the main setting and three training seeds. Flow in Table~4 of the main paper measures displacement in pixels using RAFT~\citep{raft} on consecutive-frame pairs sampled every eight frames. For each clip, we average the top-$5\%$ flow magnitude across these pairs, then report the median across clips.

\subsection{Second Model Family}
\label{app:third}
The CogVideoX-5B student is initialized from the teacher weights and trained with DMD on the same VidProM captions, eight GPUs and total batch $8$, with the critic ratio, learning rates and auxiliary weights of the main setting. Only the MF and VF references are re-estimated from 64 initialization samples. The human comparison is a separate study with 24 prompts, two noise seeds and 20 participants who passed the attention check (of 21 responses).

\subsection{Larger Batch}
\label{app:batch64}
\begin{table}[t]
\centering
\setlength{\abovecaptionskip}{0pt}
\setlength{\belowcaptionskip}{4pt}
\caption{\textbf{Total batch $64$} (single run per method, same 1K and 3K iterations as Table~2 of the main paper). dyn.: dynamic degree at 3K.}
\label{tab:batch64}
{\footnotesize
\setlength{\tabcolsep}{1pt}
\begin{tabular*}{\linewidth}{@{\extracolsep{\fill}}lccc@{}}
\toprule
& \multicolumn{2}{c}{\textbf{\vbone{}}\,$\uparrow$} & \\
\cmidrule(lr){2-3}
\textbf{Method} & 1K & 3K & \textbf{dyn.} \\
\midrule
\textsc{dmd} & $86.13$ & $80.74$ & $0.23$ \\
\methodB{} & $\mathbf{87.63}$ & $86.52$ & $0.78$ \\
\methodD{} & $87.29$ & $\mathbf{87.05}$ & $\mathbf{0.86}$ \\
\bottomrule
\end{tabular*}}
\end{table}

All other runs use a total batch of $8$. Table~\ref{tab:batch64} trains each method once with a total batch of $64$, the batch size of the published Causal Forcing recipe~\citep{causalforcing}, for the same 1K and 3K iterations. At 1K both methods are already above DMD on \vbone{}, as in the main setting. DMD still collapses at this batch size: its 3K dynamic degree is $0.23$ and its \vbone{} falls by $5.39$ points from 1K to 3K. \methodB{} and \methodD{} keep a dynamic degree of $0.78$ and $0.86$ and lose $1.11$ and $0.24$ points, so at 3K both exceed DMD on \vbone{} and dynamic degree. At 3K both methods also stay above DMD at 1K, and, as in the main setting, \methodD{} has the highest 3K \vbone{}.

\subsection{Per-seed Results}
\label{app:perseed}
\begin{table*}[!t]
\centering
\begin{minipage}{0.98\linewidth}
\centering
\small
\setlength{\abovecaptionskip}{0pt}
\setlength{\belowcaptionskip}{6pt}
\caption{\small Per-seed \vbone{} and \vbtwo{} scores underlying Table~2 of the main paper. Each row lists the three training runs in the same order (s0/r1/r2) at both checkpoints.}
\label{tab:perseed}
{
\renewcommand{\arraystretch}{1.05}
\setlength{\tabcolsep}{4pt}
\begin{tabular*}{\linewidth}{@{\extracolsep{\fill}}llcc@{}}
\toprule
\textbf{Method} & \textbf{step} & \textbf{\vbone{}} (s0/r1/r2) & \textbf{\vbtwo{}} (s0/r1/r2) \\
\midrule
\textsc{dmd} & 3000 & 81.45 / 80.81 / 80.21 & 51.87 / 50.70 / 53.36 \\
\textsc{avd} & 3000 & 81.24 / 83.96 / 78.97 & 50.26 / 51.48 / 48.62 \\
\textsc{avd-official} & 3000 & 81.95 / 84.73 / 82.18 & 53.58 / 53.14 / 47.60 \\
\methodB{} & 3000 & 86.22 / 84.66 / 84.03 & 55.04 / 52.97 / 53.79 \\
\methodD{} & 3000 & 87.20 / 87.08 / 86.31 & 53.13 / 50.36 / 52.67 \\
\midrule
\textsc{dmd} & 1000 & 85.73 / 86.72 / 86.85 & 51.39 / 48.78 / 51.03 \\
\textsc{avd} & 1000 & 81.73 / 79.98 / 83.70 & 46.03 / 48.45 / 50.03 \\
\textsc{avd-official} & 1000 & 87.76 / 84.34 / 80.42 & 52.60 / 50.35 / 46.25 \\
\methodB{} & 1000 & 86.83 / 86.70 / 87.48 & 51.31 / 52.15 / 54.60 \\
\methodD{} & 1000 & 87.08 / 87.44 / 87.13 & 53.21 / 52.54 / 51.76 \\
\bottomrule
\end{tabular*}
}
\end{minipage}
\end{table*}

Table~\ref{tab:perseed} reports the individual \vbone{} and \vbtwo{} scores for the three training runs of each method at 1K and 3K iterations. At 3K, with dynamic degree weighted as in the main paper, all three runs of \methodD{} and of \methodB{} exceed every vanilla DMD run on \vbone{}. On \vbtwo{} the runs of both methods overlap with those of DMD, and the diagnostic \vbone{} total without dynamic degree favors DMD (Table~5 of the main paper).

\section{Diagnosis and Temporal Subspace Analysis}
\label{app:evidence}

\subsection{Diagnosis details}
\label{app:diagnosis}

This section reports the full measurements summarized in Table~1 of the main paper, all taken on fixed clips or checkpoints.

\paragraph{Teacher gains.} Table~\ref{tab:a3x} reports the gain $g_P$ of the teacher's guided clean prediction for three component maps $P$: the temporal deviations, and the spatial high-pass and low-pass parts of the temporal mean. Because the gain is an alignment along the source direction, a value near one means that the component is neither contracted nor expanded along its own direction. On teacher samples, the low-pass mean keeps a gain of $0.94$--$1.00$, whereas the deviations and the high-pass mean contract increasingly with noise, the high-pass mean as much as or more than the deviations. The dispersion ratio $r$ compares the across-seed spread of $\Tmean(\xbar)$ with that of $\Tmean(\vx_0)$. Values of $0.97$--$1.07$ indicate that the predicted mean keeps the seed-to-seed variation of the clips rather than collapsing toward a prompt average. On \methodB{} samples at 1K and 3K, the low-pass mean is preserved and the deviations contract at $\sigma=0.9$ (gains $0.80$ and $0.83$), less than on teacher samples at the same noise level, and the high-pass mean contracts more ($0.58$ and $0.68$).

\begin{table}[ht]
\caption{Channel gains $g=\langle P(\xbar),P(\vx_0)\rangle/\|P(\vx_0)\|_2^2$ of the teacher's guided clean prediction at noise level $\sigma$ (means over $24$ clips of four noise draws each), for $P$ the temporal deviations (dev.), the spatial high-pass (HP) and low-pass (LP) of the temporal mean. $r$ is the across-seed dispersion of $\Tmean(\xbar)$ relative to that of $\Tmean(\vx_0)$ ($1$: equal dispersion, with a $95\%$ bootstrap interval over prompts). Gain measures alignment along the source direction, rather than reconstruction accuracy. Gains without guidance agree within $0.03$ at $\sigma\le0.9$ and are about $0.1$ lower at $\sigma=0.98$.}
\label{tab:a3x}
\begin{center}
{\scriptsize
\setlength{\tabcolsep}{4pt}
\begin{tabular}{llcccc}
\toprule
\textbf{Clips} & $\sigma$ & $g_{\mathrm{dev}}$ & $g_{\mathrm{HP}}$ & $g_{\mathrm{LP}}$ & $r$ \\
\midrule
14B teacher samples & 0.5  & 0.91 & 0.91 & 1.00 & 1.00 [0.99, 1.00] \\
                    & 0.9  & 0.72 & 0.64 & 0.99 & 0.97 [0.94, 0.98] \\
                    & 0.98 & 0.45 & 0.27 & 0.94 & 1.07 [0.96, 1.23] \\
\midrule
\methodB{} 1K       & 0.9  & 0.80 & 0.58 & 1.00 & 1.00 [0.98, 1.01] \\
\methodB{} 3K       & 0.9  & 0.83 & 0.68 & 1.01 & 1.00 [0.98, 1.01] \\
\bottomrule
\end{tabular}}
\end{center}
\end{table}

\paragraph{Teacher loss on temporal contractions.} Table~\ref{tab:a2} compares the teacher's conditional flow-matching loss on three pixel-space contractions of each clip with its loss on the clip itself. At both checkpoints and every tested noise level, both temporal contractions lower the loss, and by more than a per-frame spatial blur does. This is consistent with a teacher that does not penalize the removal of temporal variation. The blur is not matched in strength to the temporal contractions, and the values are medians over 24 \methodB{} clips per checkpoint, without intervals.

\begin{table}[ht]
\caption{The 14B teacher's conditional flow-matching loss on three pixel-space contractions of a clip, relative to its loss on the clip itself (median over clips). Guided losses agree within $0.02$. Static: every frame replaced by the first frame. Temporal mean: every frame replaced by the per-pixel temporal mean. Blur: per-frame Gaussian blur of $2$ pixels. Clips are \methodB{} samples at 1K and 3K ($24$ each).}
\label{tab:a2}
\begin{center}
{\scriptsize
\setlength{\tabcolsep}{4pt}
\begin{tabular}{llcccc}
\toprule
\textbf{Clips} & \textbf{Control} & $\sigma=0.2$ & $0.4$ & $0.6$ & $0.8$ \\
\midrule
\methodB{} 1K & static        & 0.35 & 0.53 & 0.61 & 0.65 \\
              & temporal mean & 0.29 & 0.45 & 0.56 & 0.58 \\
              & spatial blur  & 0.80 & 0.84 & 0.90 & 0.98 \\
\midrule
\methodB{} 3K & static        & 0.37 & 0.59 & 0.73 & 0.81 \\
              & temporal mean & 0.27 & 0.45 & 0.57 & 0.61 \\
              & spatial blur  & 0.77 & 0.81 & 0.88 & 0.98 \\
\bottomrule
\end{tabular}}
\end{center}
\end{table}

\paragraph{Critic and teacher alignment.} Table~\ref{tab:alignment} reports the deviation alignments $A_T$ and $A_F$ of Sec.~3.2 of the main paper. Since $\dev$ is linear, $\langle\dev(\xhat),\dev(\widehat{\vg}_{\mathrm{DMD}})\rangle=\|\dev(\xhat)\|_2^2(A_F-A_T)/n$, and an output-space step $\xhat\leftarrow\xhat-\eta\widehat{\vg}_{\mathrm{DMD}}$ gives
\[
\bigl\|\dev\bigl(\xhat-\eta\widehat{\vg}_{\mathrm{DMD}}\bigr)\bigr\|_2^2
=\|\dev(\xhat)\|_2^2\Bigl[1-\tfrac{2\eta}{n}(A_F-A_T)\Bigr]+O(\eta^2).
\]
A positive difference thus contracts latent temporal variance to first order in output space, although the actual update passes through the generator parameters. At 1K the pooled difference is positive and its interval excludes zero. At 3K the pooled interval includes zero. The three lower noise levels stay positive with intervals excluding zero, whereas the noisiest level is negative ($-6.2\pm6.0$), with an interval that is also wide at 1K, so ``vanishes at 3K'' in Sec.~3.2 of the main paper refers to the pooled value. These small differences come from one DMD run and motivate protecting the deviations with floors rather than explain the collapse.

\begin{table*}[t]
\caption{Deviation alignment of the teacher's guided clean prediction ($A_T$) and of the trained critic's clean prediction ($A_F$) with the student's own deviations, $A=\langle\dev(\vx),\dev(T)\rangle/\|\dev(\vx)\|_2^2$, on $72$ clips of the vanilla DMD run at each checkpoint, two noise draws per clip and noise level. $A_F-A_T>0$ means a descent step using the sampled DMD direction contracts latent temporal variance to first order in output space. $\pm$ gives the half-width of the $95\%$ bootstrap interval over clips, the larger side where it is asymmetric. The ``$>0$'' column reports the fraction of clips with a positive difference.}
\label{tab:alignment}
\begin{center}
{\scriptsize
\setlength{\tabcolsep}{4pt}
\begin{tabular}{lcccccc}
\toprule
\textbf{Checkpoint} & $\sigma$ & $A_T$ & $A_F$ & $A_F-A_T$ ($\times10^{-3}$) & $>0$ & \textbf{dev.\ share of update} \\
\midrule
1K & 0.556 & 0.934 & 0.937 & $3.2\pm0.3$ & 0.99 & 0.91 \\
1K & 0.769 & 0.872 & 0.876 & $3.9\pm0.8$ & 0.90 & 0.88 \\
1K & 0.882 & 0.784 & 0.789 & $5.8\pm1.7$ & 0.81 & 0.84 \\
1K & 0.952 & 0.618 & 0.620 & $1.4\pm4.6$ & 0.60 & 0.75 \\
1K & all   &       &       & $3.6\pm1.4$ & 0.76 & \\
\midrule
3K & 0.556 & 0.931 & 0.933 & $1.2\pm0.4$ & 0.81 & 0.90 \\
3K & 0.769 & 0.869 & 0.872 & $2.7\pm0.7$ & 0.86 & 0.87 \\
3K & 0.882 & 0.784 & 0.791 & $6.8\pm1.5$ & 0.89 & 0.83 \\
3K & 0.952 & 0.630 & 0.624 & $-6.2\pm6.0$ & 0.36 & 0.73 \\
3K & all   &       &       & $1.1\pm1.9$ & 0.50 & \\
\bottomrule
\end{tabular}}
\end{center}
\end{table*}

\subsection{Mechanism Probes}
\label{sec:mechanism}
Sec.~3.2 of the main paper measured what the teacher preserves. Here we edit the DMD direction itself by removing the critic.

\paragraph{Removing the critic.}
Replacing $T_\psi$ by $\xhat$ turns the DMD direction into the critic-free direction $\xhat-T_\phi$ in both subspaces. In a \methodB{} run with this change, imaging quality at 3K is $35.0$ instead of $71.3$ for a critic-equipped run, and aesthetic quality $42.3$ instead of $66.7$ (Appendix~\ref{app:axis}). \method{} keeps the critic and adds the critic-free direction only in $\mathcal S_0$. This control does not test the restriction to one subspace.

\subsection{Subspace properties}
\label{app:spectrum}

We first establish how the auxiliary objectives act on the temporal subspaces defined in Sec.~3.2 of the main paper. For video latents $\vx,\vy\in\mathbb R^{F\times C\times H\times W}$, the mean-broadcasting projector $\Pi_0$ and deviation projector $\Pi_\perp=I-\Pi_0$ satisfy
\begin{equation}
\label{eq:orthogonal}
\begin{aligned}
\vx&=\Pi_0\vx+\Pi_\perp\vx,\\
\langle\Pi_0\vx,\Pi_\perp\vy\rangle&=0,\\
\|\vx\|_2^2&=F\|\Tmean(\vx)\|_2^2+\|\dev(\vx)\|_2^2.
\end{aligned}
\end{equation}
Here $F$ is the number of latent frames, $C$ the number of latent channels, and $H,W$ the latent height and width. The operator $I$ is the identity, $\Tmean(\vx)$ is the temporal mean, and $\dev(\vx)$ is the zero-mean deviation sequence. The factor $F$ accounts for repeating the mean across all frames. All norms and inner products here are Frobenius. The same decomposition applies to the sampled DMD direction in Eq.~(1) of the main paper:
\begin{equation}
\label{eq:grad-split}
\widehat{\vg}_{\mathrm{DMD}}
=\Pi_0\widehat{\vg}_{\mathrm{DMD}}
+\Pi_\perp\widehat{\vg}_{\mathrm{DMD}}.
\end{equation}

\paragraph{Gradients of the auxiliary objectives.}
For the student sample $\xhat$ and detached teacher prediction $\xbar$, IA and the active global MF term have output gradients
\begin{equation}
\begin{aligned}
\nabla_{\xhat}\mathcal L_{\mathrm{IA}}
&=\frac{2}{FCHW}\Pi_0(\xhat-\xbar),\\
\nabla_{\xhat}\mathcal L_{\mathrm{MF}}
&=-\frac{2\dev(\xhat)}{FCHW\bigl(V(\xhat)+\varepsilon\bigr)}.
\end{aligned}
\end{equation}
The first lies in the time-constant subspace $\mathcal S_0$, while the second lies in the zero-mean subspace $\mathcal S_\perp$. The variance $V$ and stabilizer $\varepsilon$ are those of MF (Sec.~3.4 and Eq.~(6) of the main paper). An output-space descent step on active MF expands existing deviations, but its gradient is zero at an exactly time-constant video.

The same subspace property extends to MSW, VF, and TS. Hold the teacher-derived weights and warm-up references fixed, as in the implementation, and let $\mathbf c\in\mathcal S_0$ be any offset shared by all frames. Then $\dev(\xhat+\mathbf c)=\dev(\xhat)$, so global and saliency-weighted MF and VF are unchanged. Since the temporal blur $\mathcal B$ preserves time-constant signals, $(I-\mathcal B)\mathbf c=0$, and TS is also unchanged. Thus each of these latent objectives $\mathcal J$ satisfies
\begin{equation}
\mathcal J(\xhat+\mathbf c)=\mathcal J(\xhat),
\qquad
\Pi_0\nabla_{\xhat}\mathcal J=\mathbf 0.
\end{equation}
The gradient identity follows by differentiating along every direction in $\mathcal S_0$ and also holds for the subgradients used at hinges and absolute values. These objectives therefore have output gradients orthogonal to IA. This orthogonality holds for the output-space gradients of the auxiliary terms only. The DMD gradient has components in both subspaces (Eq.~(\ref{eq:grad-split})), and all terms reach the generator parameters through the same Jacobian, so their parameter-space updates are not orthogonal in general.

\paragraph{DMD with IA in the two subspaces.}
Let $N=FCHW$ and $\vg=T_\psi-T_\phi$. Since $\xbar$ is the detached $T_\phi$, the DMD and IA output gradients are $\vg/(nN)$ and $(2/N)\Pi_0(\xhat-\xbar)$, where the per-sample normalizer $n$ of Eq.~(1) of the main paper is detached. Hence
\begin{equation}
\label{eq:ia-residual-derivation}
nN\,\nabla_{\xhat}\bigl(\mathcal L_{\mathrm{DMD}}+\lambda_{\mathrm{IA}}\mathcal L_{\mathrm{IA}}\bigr)
=\Pi_\perp\vg+\Pi_0\bigl(\vg+\lambda'(\xhat-\xbar)\bigr),
\end{equation}
with $\lambda'=2n\lambda_{\mathrm{IA}}$, which is Eq.~(5) of the main paper. Replacing $T_\psi$ by $\xhat$ in the surrogate of Sec.~3.1 of the main paper gives the critic-free output gradient
\begin{equation}
\label{eq:cf-residual}
\vg_{\mathrm{CF}}=\frac{\xhat-\xbar}{nN},
\qquad
\Pi_0\vg_{\mathrm{CF}}=\frac{1}{2n}\nabla_{\xhat}\mathcal L_{\mathrm{IA}},
\end{equation}
the clean-prediction form of score distillation~\citep{dreamfusion}, which acts in both subspaces. DMD with IA uses this direction only in $\mathcal S_0$ and keeps the critic-based DMD direction in both.

\paragraph{Variation and motion.}
Any permutation of the frames leaves $V$ unchanged, even when it disrupts a coherent action. A variance floor therefore cannot by itself enforce temporal order or plausible motion. VF measures ordered changes after temporal smoothing, while TS penalizes the fluctuations removed by that smoothing. These latent statistics still do not directly measure displacement in decoded frames.

Keeping the DMD term unchanged does not preserve every stationary point of the combined objective. IA, active floors, and TS can all contribute additional gradients. The floors become inactive when their references are met, but they remain soft penalties rather than hard constraints.

\subsection{Critic-free control}
\label{app:axis}
The coupled student and critic updates in DMD pose an optimization challenge analogous to GAN training. We examine whether removing the critic offers a viable alternative. In one \methodB{} run on the second cluster, we replace $T_\psi$ with the student's sample $\xhat$, disable critic training, and retain the auxiliary losses. By Eq.~(\ref{eq:cf-residual}), this replaces the DMD direction by the critic-free direction in both subspaces, also with IA retained, so the critic cancels from the update. Table~\ref{tab:critic-free} compares this control with a critic-equipped reference run at 1K and 3K. At both checkpoints, imaging and aesthetic quality are substantially lower than in the reference run. These single-run results support retaining the critic in the DMD update, but they do not test subtraction restricted to one subspace.

\begingroup
\setlength{\intextsep}{4pt plus 1pt minus 1pt}
\begin{table}[!htbp]
\centering
\begin{minipage}{0.98\linewidth}
\centering
\small
\setlength{\abovecaptionskip}{0pt}
\setlength{\belowcaptionskip}{6pt}
\caption{\small \textbf{Critic-free control on \methodB{}.} Imaging and aesthetic quality are evaluated with \vbone{} on the official prompts and reported on a $0$ to $100$ scale. Results use one training run per configuration on the second cluster.}
\label{tab:critic-free}
{
\renewcommand{\arraystretch}{1.05}
\setlength{\tabcolsep}{4pt}
\begin{tabular*}{\linewidth}{@{\extracolsep{\fill}}lcccc@{}}
\toprule
& \multicolumn{2}{c}{\textbf{Imaging quality}\,$\uparrow$}
& \multicolumn{2}{c}{\textbf{Aesthetic quality}\,$\uparrow$} \\
\cmidrule(lr){2-3}\cmidrule(l){4-5}
\textbf{Checkpoint} & With critic & Critic-free & With critic & Critic-free \\
\midrule
1K & 69.5 & 28.1 & 66.0 & 37.1 \\
3K & 71.3 & 35.0 & 66.7 & 42.3 \\
\bottomrule
\end{tabular*}
}
\end{minipage}
\end{table}

\endgroup

\section{Continued Distillation}
\label{app:continued-distillation}

Figure~2 of the main paper compares vanilla DMD at 1K and 3K training iterations: the 3K samples show finer details, but the motion becomes less apparent and physically implausible. VBench-1 and VBench-2 do not fully capture such changes in visual fidelity, action coherence, and object interactions over time. The benchmark scores serve as complementary references and should be interpreted alongside the visual comparisons.

\begin{figure*}[!t]
\centering
\includegraphics[width=\linewidth,height=0.9\textheight,keepaspectratio]{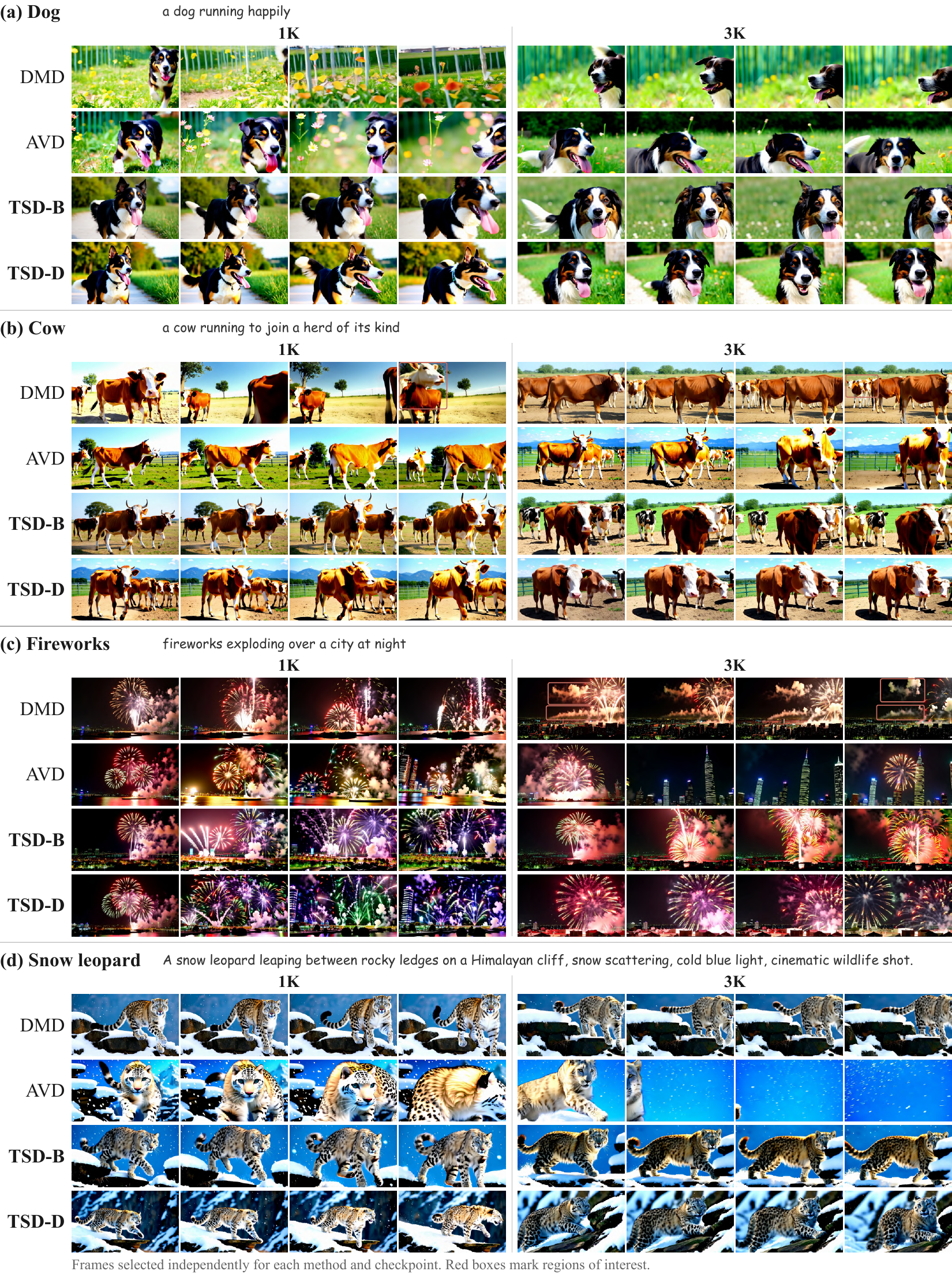}
\caption{\textbf{Continued distillation across methods.} DMD, AVD, \methodB{}, and \methodD{} at 1K (left) and 3K (right). Red boxes highlight head-body overlap and persistent smoke bands.}
\label{fig:continued-distillation}
\end{figure*}

Figure~\ref{fig:continued-distillation} compares 1K and 3K checkpoints across four scenes. Dynamic degree measures only the presence of motion above a flow threshold, and flow magnitude measures displacement rather than action coherence or physical plausibility.

The 1K DMD in Table~2 of the main paper is an early-checkpoint reference with a competitive \vbone{} score. In the user study (Appendix~\ref{app:user-study}), participants prefer \methodD{} at 3K over its 1K checkpoint on visual quality and motion (73\% and 72\% of votes, both significant after Holm correction). The automatic metrics do not show this gain, since the \vbone{} and \vbtwo{} of \methodD{} fall from 1K to 3K ($87.22\to86.86$ and $52.50\to52.05$), and the study covers only the 400 student updates after 1K at a total batch of $8$.

\subsection{Teaser Prompts}
\label{app:teaser-prompts}
The clips in Figure~1 of the main paper use the following prompts, clockwise from the jellyfish. The snow leopard comes from \methodB{} and all other clips from \methodD{}. The prompt labels in the figure are the opening words of each prompt.
\begin{itemize}[leftmargin=1.4em,itemsep=1pt,topsep=1pt]
\item A jellyfish drifting through deep blue water with its translucent bell pulsing, bioluminescent tendrils trailing behind, underwater macro cinematography.
\item fireworks exploding over a city at night
\item A close-up of an elderly man's face as he smiles and looks up from a book, wrinkles and gray beard in detail, warm library lighting, gentle camera push-in.
\item a cow running to join a herd of its kind
\item A snow leopard leaping between rocky ledges on a Himalayan cliff, snow scattering, cold blue light, cinematic wildlife shot.
\item a dog running happily
\end{itemize}

\section{User Study}
\label{app:user-study}

This section details the user study in Table~3 of the main paper. Nine comparisons share one pool of participants: \methodD{}@3K against DMD@1K, DMD@3K, \textsc{avd-official}@3K, \textsc{dmd}$+$motion GAN@3K, \methodD{}@1K, and the 50-step 14B teacher, and \methodB{}@3K against three ablations at 3K, without IA (the floors-only ablation), without VF and TS (the IA$+$MF ablation), and with the full anchor of Table~5 of the main paper in place of the mean anchor. The design, the comparisons, the criterion tested for each ablation, and the analysis were fixed before data collection.

\paragraph{Prompts and videos.}
We sampled 64 prompts from \vbone{} with a fixed random seed: 32 from the dynamic-degree list and 32 from the aesthetic and imaging quality list, after removing the prompts of an earlier pilot study (below). Each prompt is used with two noise seeds, giving 128 items. Item $(q,s)$ uses the checkpoints of training seed $(q+s)\bmod 3$, so each training seed covers 42--43 items, and both videos of a pair share the prompt, noise seed, and training seed. Students use the inference settings of Table~2 of the main paper and the teacher uses 50 steps. Each video was generated once, without selection.

\paragraph{Questions.}
Each question shows the prompt and one clip with the two videos side by side under the labels \emph{Left} and \emph{Right}, played twice ($10$\,s) and replayable. Participants rate two criteria, each with the options \emph{Left is better}, \emph{About the same}, and \emph{Right is better}. \emph{Visual quality} asks which video is sharper and more realistic, counting blur, noise, artifacts, flicker, and over-sharpening against a video, and states that how much or how naturally things move should not affect the answer. \emph{Motion} asks which video's motion looks more natural for the description, counting missing or very weak motion, motion that is faster, larger, or more erratic than plausible, deformation, and jitter against a video, and asks participants not to prefer a video just because it moves more. Camera movement counts only when it suits the scene. Method names, checkpoints, and training steps are never shown.

\paragraph{Assignment.}
The design has 15 slots, one for each comparison with \methodD{}@3K and three for each ablation, and every slot rates each of the 128 items exactly once. Each of the 64 questionnaire versions contains two pairs per slot, one from each prompt stratum and with the reference on the left in one of them, so a version holds 30 pairs on 30 distinct prompts, and with three attention checks 33 questions in total. Each ablation item is therefore rated by three different participants, not always with the reference on the same side. Each version has its own fixed random question order, and each participant completes one version.

\paragraph{Attention checks and screening.}
The three attention checks sit at random positions within questions 5--10, 15--20, and 25--30 of the 33. They pair a video of the shared initialization on a separate prompt with a degraded copy: blurred by $4\times$ down- and up-sampling, with its frames permuted so that neighboring frames are at least three frames apart in the original order, or both. A response passes if the intact video is preferred on visual quality, on motion, or on both, respectively. We exclude responses that fail a check or give the same answer to all 60 ratings. All 64 responses passed both screens, and their respondent identifiers are distinct. The survey export does not record completion time, so a planned minimum-time screen could not be applied.

\paragraph{Analysis.}
Each rating is counted from the perspective of the reference as a win, tie, or loss. For each comparison and criterion, we test whether the reference receives 50\% of the non-tie votes with a score test whose variance is two-way cluster-robust over participants and prompts and floored at the binomial variance under the null. The 95\% interval inverts the same test, so the interval and the test always agree. Holm correction runs over two families: the ten cells of the five comparisons of \methodD{}@3K with baselines and its 1K checkpoint, and the three pre-specified cells of the ablations, visual quality for the floors-only ablation and motion for the other two. The other criterion of each ablation is exploratory, and the teacher comparison is a descriptive reference, so these cells report intervals only.

\begin{table}[t]
\centering
\setlength{\abovecaptionskip}{0pt}
\setlength{\belowcaptionskip}{4pt}
\caption{User-study statistics behind Table~3 of the main paper. Votes: W/T/L counts for the reference. Win: \% of non-tie votes for the reference, with a 95\% interval from the inverted score test (two-way cluster-robust over participants and prompts). $p$: Holm-corrected, within each block. ``--'': not tested (teacher reference and exploratory ablation criteria). Prompts: number of the 64 prompts on which the reference receives more / fewer votes.}
\label{tab:user-study-stats}
{\scriptsize
\setlength{\tabcolsep}{1.3pt}
\begin{tabular}{@{}llcccc@{}}
\toprule
\textbf{vs.} & \textbf{Crit.} & \textbf{Votes} & \textbf{Win [95\% CI]} & $p$ & \textbf{Prompts} \\
\midrule
\multicolumn{6}{@{}l}{\methodD{}@3K vs.} \\
\textsc{dmd}@1K            & Visual & 114/4/10 & 91.9 [85.8, 96.5]  & $<$0.001 & 54/1 \\
                           & Motion & 115/8/5  & 95.8 [90.6, 99.4]  & $<$0.001 & 59/0 \\
\textsc{dmd}@3K            & Visual & 112/1/15 & 88.2 [81.4, 94.6]  & $<$0.001 & 51/3 \\
                           & Motion & 122/0/6  & 95.3 [90.2, 99.6]  & $<$0.001 & 58/0 \\
\textsc{avd-official}@3K   & Visual & 125/1/2  & 98.4 [94.4, 100.0] & $<$0.001 & 62/0 \\
                           & Motion & 126/0/2  & 98.4 [94.5, 100.0] & $<$0.001 & 62/0 \\
\textsc{dmd}+motion GAN@3K & Visual & 110/2/16 & 87.3 [80.4, 93.8]  & $<$0.001 & 49/1 \\
                           & Motion & 111/1/16 & 87.4 [80.5, 93.5]  & $<$0.001 & 48/1 \\
\methodD{}@1K              & Visual & 94/0/34  & 73.4 [64.5, 82.4]  & $<$0.001 & 37/7 \\
                           & Motion & 92/5/31  & 74.8 [66.2, 83.4]  & $<$0.001 & 40/7 \\
14B teacher                & Visual & 50/17/61 & 45.0 [34.1, 55.9]  & --       & 18/27 \\
                           & Motion & 57/20/51 & 52.8 [42.3, 63.3]  & --       & 21/17 \\
\midrule
\multicolumn{6}{@{}l}{\methodB{}@3K vs.} \\
w/o IA                     & Visual & 291/11/82  & 78.0 [71.3, 84.7] & $<$0.001 & 51/6 \\
                           & Motion & 293/10/81  & 78.3 [71.8, 84.9] & --       & 54/7 \\
w/o VF, TS                 & Visual & 282/14/88  & 76.2 [70.5, 82.0] & --       & 54/6 \\
                           & Motion & 297/9/78   & 79.2 [73.7, 84.7] & $<$0.001 & 54/4 \\
Full anchor                & Visual & 137/116/131 & 51.1 [43.2, 59.1] & --      & 29/28 \\
                           & Motion & 199/108/77 & 72.1 [66.3, 77.9] & $<$0.001 & 51/10 \\
\bottomrule
\end{tabular}}
\end{table}

\paragraph{Results.}
Table~\ref{tab:user-study-stats} reports the statistics behind Table~3 of the main paper. Percentages in this paragraph are of non-tie votes, whereas Table~3 of the main paper reports shares of all votes, including ties. All 13 tested preferences are significant after Holm correction. Against DMD at either checkpoint and against \textsc{avd-official}, \methodD{}@3K receives at least 88.2\% of the non-tie votes and fewer votes than the alternative on at most 3 of the 64 prompts. The comparison with \methodD{}@1K is the closest of the tested comparisons with \methodD{}@3K ($73.4\%$ and $74.8\%$). Against the teacher, \methodD{}@3K receives fewer votes than the teacher on 27 prompts for visual quality and 17 for motion. Counting ties as half a vote keeps every tested preference at or above $65.9\%$, and split by training seed, every tested preference is at least $67\%$ in each of the three seeds. Split by prompt stratum, every tested preference receives 69--100\% of the non-tie votes on motion prompts and 71--98\% on quality prompts, so neither half drives the results. Position bias is small: the left video receives $50.3\%$ of the non-tie votes. Agreement between the two criteria is lowest where the compared methods differ mainly in motion: only 35\% of the full-anchor questions receive the same verdict on both, against 67--96\% in the comparisons of \methodD{}@3K with baselines and its 1K checkpoint, where both criteria favor \methodD{}@3K. Participants therefore judged the criteria separately rather than carrying motion into visual quality.

\paragraph{Pilot study.}
An earlier pilot with 93 participants rated 16 other \vbone{} prompts, one noise seed each, with a motion question that counted only too little motion against a video and with \methodD{}@3K as the reference in every comparison. Its preferences for \methodD{}@3K over DMD at both checkpoints, \textsc{avd-official}, \textsc{dmd}$+$motion GAN, and \methodD{}@1K point in the same direction as in this study ($70.6$--$100\%$ of non-tie votes). It also compared \methodD{}@3K with \methodB{}@3K directly and found no significant preference ($56.0\%$ and $55.6\%$ of non-tie votes, Holm $p=0.45$, with 31\% and 26\% ties), so the pilot detects no perceptual effect of MSW and CRS. Pilot votes are not pooled with this study.

\paragraph{Scope.}
Each version has one participant, so each comparison of \methodD{}@3K rests on 64 participants with two votes each. Intervals account for variation across participants and prompts, and the three training seeds enter only through the item assignment. The ablations use \methodB{} as the reference, and this study compares \methodD{} with neither \methodB{} nor \textsc{avd-full}, as its only AVD variant is \textsc{avd-official}, without AVD's regression term. The comparison between \methodD{} and \methodB{} in the Limitations therefore rests on the pilot alone. In the descriptive comparison with the teacher (dynamic degree $0.69$), \methodD{}@3K receives $53\%$ of non-tie motion votes under a question that counts implausibly fast, large, or erratic motion against a video, so participants did not favor the teacher's lower dynamic degree, but only $45\%$ on visual quality, with both intervals including $50\%$.

\end{document}